# GaussiGAN: Controllable Image Synthesis with 3D Gaussians from Unposed Silhouettes

Youssef A. Mejjati[1], Isa Milefchik[5], Aaron Gokaslan[2], Oliver Wang[3], Kwang In Kim[4], James Tompkin[5]
[1]University of Bath, [2]Cornell University, [3]Adobe, [4]UNIST, [5]Brown University

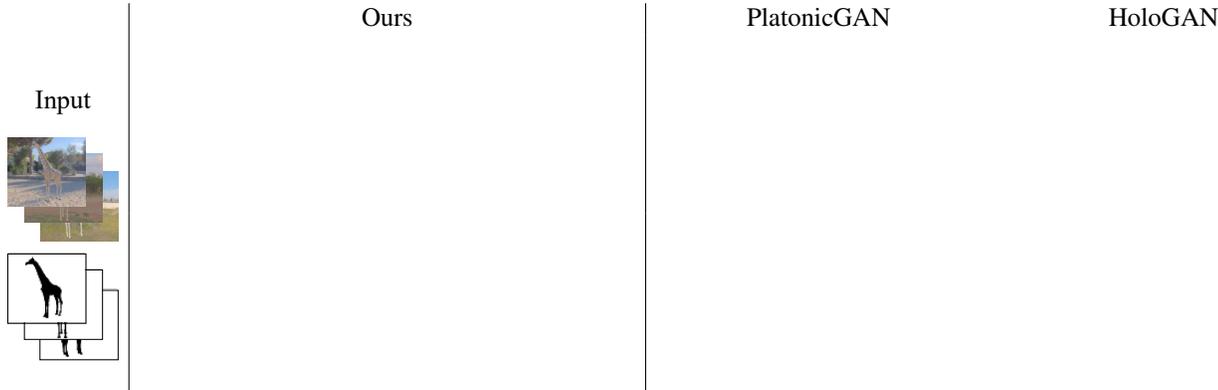

Figure 1. From silhouettes of an object with unknown camera and object pose, we infer a set of 3D Gaussians to represent coarse shape and pose. This allows 'rigging' detailed 2D mask generation and associated detailed 2D texture, disentangling of texture variation (here illumination), and interactive camera and pose control for object insertion onto backgrounds. PlatonicGAN [6] (*mid right*) generates inconsistent shapes from silhouettes, and texturing is a challenge. HoloGAN [33] (*far right*) on silhouettes has an inconsistent 3D space and, without explicit shape, rotation and lighting are entangled on RGB foregrounds. *Please view in Adobe Acrobat to see animations.*

## Abstract

*We present an algorithm that learns a coarse 3D representation of objects from unposed multi-view 2D mask supervision, then uses it to generate detailed mask and image texture. In contrast to existing voxel-based methods for unposed object reconstruction, our approach learns to represent the generated shape and pose with a set of self-supervised canonical 3D anisotropic Gaussians via a perspective camera, and a set of per-image transforms. We show that this approach can robustly estimate a 3D space for the camera and object, while recent baselines sometimes struggle to reconstruct coherent 3D spaces in this setting. We show results on synthetic datasets with realistic lighting, and demonstrate object insertion with interactive posing. With our work, we help move towards structured representations that handle more real-world variation in learning-based object reconstruction.*
*Project webpage:* visual.cs.brown.edu/gaussigan

## 1. Introduction

Inferring poseable 3D object representations from image data for tasks like controllable generation is complex: objects are visible under unknown perspective 3D cameras, have different forms and shapes, have moving parts in different poses, and vary appearance due to lighting (among other variations). Oftentimes, we have little supervision for learning parametrized models for these properties. Yet, for control, we would like to discover a flexible structure that does not presume an underlying template. This task is similar to discovering an 'artist's mannequin' for an object.

We consider a setting with only mask supervision. This problem is related to shape from silhouette, which assumes known camera poses and rigid objects and recovers shape via intermediate voxels. However, in our setting, we do not know the camera poses and the articulated object in the dataset has unknown variable pose per image. Prior approaches to this problem domain rely on voxel prediction [6], which can conflate 3D spaces across images in the unposed camera setting (Fig. 1, center right), and a deep voxel-based representation [33] which provides too much freedom to recover coherent 3D spaces (Fig. 1, right).

Instead of voxels, we aim to estimate object 'parts' in a robust way; this lets us abstract the structure from the detail in this challenging inference case. For this, we introduce a mixture of anisotropic 3D Gaussians as a coarse implicit geometry proxy. These are low dimensional to infer, have an analytically-differentiable projection model under perspective cameras, are composable for parts, can represent position, scale, and rotation to model part pose transforms, and are simple to self-supervise. Even then, inferring a coherent 3D space across images is tricky. For this, we employ a canonical 3D Gaussian set plus per-image camera and per-Gaussian transformation parameters that describe camera and object pose for each image. Through training via 2D silhouette reconstruction, our representation and losses associate object parts with Gaussians, despite not having any part-level supervision.

For evaluation, we control input variation using synthetic data with varying camera pose, object pose, and illumination, from which to show recovery of this low-dimensional structure. Then, using the learned Gaussians within 2D RGB generation, we show disentangling of pose, view-dependent texture, and shading variation caused by lighting differences. This lets us insert objects at arbitrary viewing angles into backgrounds, and interactively adjust object pose by directly manipulating the Gaussians of our 'artist's mannequin'. Looking forward, our work implies a structure to robustly handle pose and shape to better cope with the increased variation in 'in the wild' datasets.

## 2. Related work

**Image and object generation and insertion.** GANs have shown tremendous progress in learning-based whole image generation [5, 55, 57], including disentangling latent features [38, 23, 14]. Beyond whole images, research has investigated how to learn to generate and add 2D objects to a given background [51], including 2D object shape generation [21] also via bounding boxes [53], completing bounding boxes with texture [7], learning to warp 2D foregrounds [26], insert 2D objects [37], or animate in 2D [44]. We learn an explicit 3D representation that allows controllable image generation.

**Unsupervised keypoint and part detection.** Gaussians are related to keypoints and parts. Learning these is possible with supervision [27, 39, 32] and without. Here, Thewlis *et al*. [49, 48] use equivariance under 2D image transformations like warping to predict object keypoints; however, this requires the transformations to be known. To address this, Jakab *et al*. [8] learn keypoints in a self supervised way by reconstructing an object's appearance and geometry from different viewpoints. To make these intuitive, Jakab *et al*. later use a skeleton prior (*e.g*., face, eyes, nose) to guide a discriminator [9]. This has been extended to video prediction with realistic motion [16]. Some methods use Gaussians within their pipelines. Lorenz et al. [28] predict unconstrained 2D activation maps per part for unsupervised part discovery, then estimate 2D Gaussian parameters from these to mark keypoints. Instead, we directly learn a set of 3D Gaussians to describe the shape and pose of an object.

**3D object representations.** Learned representations exist for taking 3D input data like point clouds [1], volumes [43], or meshes [52, 12, 3] and generating 3D output data. These include techniques to fit sets of Gaussians to 3D shapes using 3D supervision [4], and by combining 3D supervision with multi-view silhouette losses [54]. Some works use predefined detailed canonical 3D meshes for 2D images [59], e.g., to learn surface parameterizations [20, 31]. Other works learn representations from 2D input data via 3D representations [35, 15] and flow [40], and often need multi-view camera information given at training time [36, 29]. For instance, DeepVoxels [45] projects RGB values on known camera rays to learn a deep voxel space that reproduces 2D inputs when projected and decoded. Other works require object-specific pose information, such as human skeletal data [19]. Without camera poses, Lei et al. build surface parametrizations for rigid 3D objects [22].

For image generation, few works take only 2D input and *no* camera or object pose information for supervision— this is hard as there is no explicit constraint on the 3D space. Schwarz et al. and Niemeyer and Geiger generate fields for 3D objects [43, 34]. Liao et al. use cube and sphere mesh proxies to represent multiple simple scene objects [24]. HoloGAN uses deep voxels within an implicit rotation space [33], and PlatonicGAN uses discrimination on random rotations to learn a generative voxel space [6]. Different geometry and appearance proxies have trade-offs, e.g., voxels can capture shape detail but are a high dimensional space to predict; our 3D Gaussian proxy is low dimensional but coarse and can capture transformable parts.

## 3. Learning Gaussian proxies for shape & pose

We wish to reconstruct parts-based models for objects as a set of Gaussian proxies. To accomplish this, we will use supervision only via performing the task of mask reconstruction. We train a network to predict a set of 3D anisotropic Gaussians as coarse proxies for the objects' shape and pose, where each Gaussian emerges to loosely represent one part of the object; the mean defines the position and its covariance defines the rotation and scale of the part. Prediction is trained by projecting Gaussians into a perspective camera and transforming them into a detailed mask via a GAN. In this process, we recover a canonical Gaussian representation for the object, from which specific pose and shape transforms are estimated per image.

**Input masks and anisotropic 3D Gaussians.** We start with a dataset of 256×256 binary segmentation masks $m \in \mathcal{M}$ of an object under varying unknown camera parameters and object poses. We also require a given number $K$ of unnormalized anisotropic 3D Gaussians $\{\mathcal{G}_k\}_{k=1}^{K}$ (Fig. 2). Each Gaussian $\mathcal{G}_k$ has mean vector $\boldsymbol{\mu}_k \in \mathbb{R}^3$ and covariance matrix $\boldsymbol{\Sigma}_k \in \mathbb{R}^{3\times 3}$ with its density declared as:

$$\mathcal{G}_k(\mathbf{x}) = \exp\left(-(\mathbf{x}-\boldsymbol{\mu}_k)^\top \boldsymbol{\Sigma}_k^{-1}(\mathbf{x}-\boldsymbol{\mu}_k)\right). \quad (1)$$

**Camera.** We declare a general perspective pinhole camera with intrinsic matrix $\mathbf{K}$, rotation $\mathbf{R}$, and translation $\mathbf{t}$ such that camera matrix $\mathbf{P}$ is represented as $\mathbf{K}[\mathbf{R}, \mathbf{t}]$. To project a 3D anisotropic Gaussian into our camera's image plane to produce a 2D anisotropic Gaussian, we use analytically-differentiable projection function $\pi$ [46]. This is valid for perspective cameras, unlike orthographic [6] or para-perspective [54] projection models that are less applicable to real-world cameras. Please see supplemental material for details of $\pi$. In our experiments, $\mathbf{K}$ is fixed across images and approximately matches that in the data.

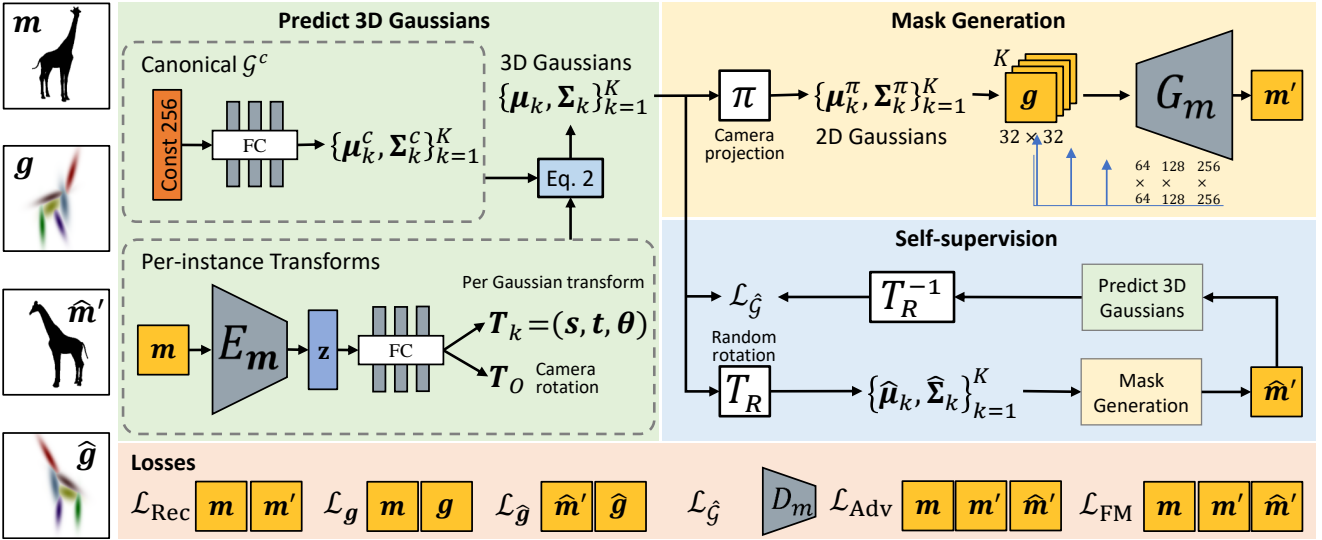

Figure 2. Learning a $K$-part 3D Gaussian representation with only mask supervision $m$. *Green:* For each instance, we predict 3D anisotropic Gaussians by combining a canonical representation with scale, rotation, and translation transforms. *Yellow:* We project these down to 2D Gaussians in an analytically-differentiable way, then sample these into $K$ maps. $g$ conditions network $G_m$ to generate a detailed mask $m'$ as a reconstruction of $m$. *Blue:* To learn a meaningful and smooth 3D space, we self supervise reconstruction by forcing a random rotation of our estimated 3D Gaussians to also produce a plausible mask $\widehat{m}'$ *and* for its 3D Gaussian prediction to be consistent after the inverse rotation. *Orange:* We penalize reconstruction losses on masks and promote realism via adversarial discrimination.

**Canonical Gaussians.** Given a 256-dimensional constant [13] as input, we use a fully connected network $E_{\mathcal{G}^c}$ to predict the canonical 3D Gaussians $\mathcal{G}_k^c$ each parameterized by a mean and covariance $(\boldsymbol{\mu}^c, \boldsymbol{\Sigma}^c)$ (Fig. 2, green, top).

**Per-image Gaussian transforms.** Given an input mask $m$, we extract a latent vector representing pose $\mathbf{z} \in \mathbb{R}^8$ via a convolutional encoder network $E_m$. Then, from $\mathbf{z}$, we use a fully connected network to predict two transformations: 1) A camera transformation $\mathbf{T}_O$ that moves the camera with respect to the canonical model; in our experiments, we consider a yaw rotation $\mathbf{R}_\phi$. 2) $K$ Gaussian local transformations $\mathbf{T}_k$ consisting of scale, translation, and rotation $(\mathbf{s}_k, \mathbf{t}_k, \boldsymbol{\theta}_k)$ with each in $\mathbb{R}^3$ (Fig. 2, green, bottom).

Given the canonical parameters $(\boldsymbol{\mu}_k^c, \boldsymbol{\Sigma}_k^c)$, we obtain the per-image Gaussians $\mathcal{G}_k$ with parameters $(\boldsymbol{\mu}_k, \boldsymbol{\Sigma}_k)$ via:

$$\boldsymbol{\mu}_k = \mathbf{R}_\phi(\boldsymbol{\mu}_k^c + \mathbf{t}_k)$$
$$\boldsymbol{\Sigma}_k = (\mathbf{R}_\phi \mathbf{R}_{\theta_k} \mathbf{U}_k \mathbf{s}_k \mathbf{S}_k)(\mathbf{R}_\phi \mathbf{R}_{\theta_k} \mathbf{U}_k \mathbf{s}_k \mathbf{S}_k)^\top, \quad (2)$$

where $\mathbf{R}_{\theta_k}$ is the rotation matrix form of $\boldsymbol{\theta}_k$, and $\mathbf{S}_k$ and $\mathbf{U}_k$ are obtained via eigenvalue decomposition of $\boldsymbol{\Sigma}_k^c$: $\boldsymbol{\Sigma}_k^c =$ $(\mathbf{U}_k \mathbf{S}_k)(\mathbf{U}_k \mathbf{S}_k)^\top$. $\mathbf{S}_k$ is a diagonal matrix. The square of its $(j, j)$-th entry represents the $j$-th eigenvalue of $\boldsymbol{\Sigma}_k$. This allows us to control the scale and rotation of each individual Gaussian via the matrices $\mathbf{U}_k$ and $\mathbf{S}_k$.

**Estimating $\boldsymbol{\Sigma}$.** Training covariance $\boldsymbol{\Sigma}$ to be positive definite can be tricky; we describe and compare our eigendecomposition approach to other methods in supplemental.

**Conditional mask synthesis.** Even a large number of Gaussian proxies will not reconstruct fine mask detail. As such, we use a conditional mask generator $G_m$ to add back the detail using up-sampling transposed convolutions (Fig. 2, yellow). Given the 3D Gaussians for an image, we project them to 2D Gaussians on the image plane of our camera: $\pi(\mathcal{G}_k) = (\boldsymbol{\mu}_k^\pi, \boldsymbol{\Sigma}_k^\pi)$. Then, using the 2D version of Eq. 1, we sample the density of each projected Gaussian on a raster grid to create $K$ Gaussian maps $\{g_k\}_{k=1}^K$. These are input to $G_m$ to condition the synthesis of predicted mask $m'$, which is the learned reconstruction of $m$. We enforce a stronger effect in $G_m$ by using layer-wise conditioning via Gaussian maps at $32^2, 64^2, 128^2$, and $256^2$ resolutions.

### 3.1. Losses

We encourage our network to reconstruct an object using multiple losses, with overall energy to minimize given by:

$$\mathcal{L}(E_{\mathcal{G}^c}, E_m, G_m, D_m) = \lambda_1 \mathcal{L}_{\text{Rec}} + \lambda_2 \mathcal{L}_g +$$
$$\lambda_3 \mathcal{L}_{\widehat{\mathcal{G}}} + \lambda_4 \mathcal{L}_{\widehat{g}} + \lambda_5 \mathcal{L}_{\text{Adv}} + \lambda_6 \mathcal{L}_{\text{FM}} \quad (3)$$

**Reconstruction loss.** We encourage synthesized mask $m'$ to reconstruct input mask $m$ with an $L_1$ loss: $\mathcal{L}_{\text{Rec}}(m, m') = \|m - m'\|_1$.

| | a) Ours | b) No $\mathcal{L}_{Rec}$ | c) No $\mathcal{L}_g$ | d) No $\mathcal{L}_{\widehat{\mathcal{G}}}$ | e) No $\mathcal{G}^c$ | f) Free $\mathbf{T}_k$ |
|---|---|---|---|---|---|---|
| IoU ▲ | 83.96 | 65.09 | 82.98 | 84.75 | 86.62 | 73.16 |
| DSSIM ▼ | 6.22 | 14.20 | 6.83 | 6.16 | 5.32 | 10.94 |

Figure 3. Ablations for *Giraffe*. Note: Input masks vary per column as certain effects are only visible at particular angles; Gaussian colors vary across columns. **(a)** Our full loss model. **(b)** Without a reconstruction loss on $m'$, the Gaussians only approximately correspond to the input mask. **(c)** Without a density loss on $g$, the Gaussians do not well represent the input mask, yet $G_m$ still produces the correct mask from these less 'coherent' Gaussians. **(d)** Not 'closing the loop' in the self-supervised loss hurts self occlusion cases or when the 2D Gaussian layouts are not sufficient to recover 3D information. **(e)** Not using a canonical representation fails to rotate Gaussians recovered for thin front/back views. **(f)** Not reasonably bounding the per-instance transforms allows nonsense canonicals.

*Table:* Over the test set, mean IoU×100 and DSSIM×100 of reconstructed masks vs. ground truth masks at specific camera angles. Our qualitative results show these metrics do not tell the whole story.

**Density loss.** Even though they cannot represent fine detail in $m$, we still wish for all projected Gaussians to 1) cover regions of the mask without overlap, and 2) cover as much of the mask as possible. We encourage this via:

$$\mathcal{L}_g(m, g) = \left\| m - \Sigma_{k=1}^{K} g_k \right\|_1. \quad (4)$$

The sum over sampled 2D Gaussians is equivalent to a grayscale version of the colored parts visualization in Figure 3. Here, both inputs are in the range $[0, 1]$, and we take $g$ at our mask resolution of 256×256.

**Self-supervised transform mask loss.** We wish for the 3D space expressed through our recovered object Gaussians and camera transform parameters in $\mathbf{T}_O$ to be consistent across varying camera views even though we only have mask supervision. Thus, we randomly sample a 3D transformation $\mathbf{T}_R$, again mainly as a yaw rotation, and apply it via Eq. 2 to produce rotated 3D Gaussians $\widehat{\mathcal{G}} = (\widehat{\boldsymbol{\mu}}, \widehat{\boldsymbol{\Sigma}})$. As before, these are then projected via $\pi$ to 2D parameters $(\widehat{\boldsymbol{\mu}}^\pi, \widehat{\boldsymbol{\Sigma}}^\pi)$, then sampled into 2D maps $\widehat{g}$, and finally via $G_m$ to generate a mask $\widehat{m}'$ (Fig. 2, blue).

As $\widehat{m}'$ does not correspond to a known input image, we cannot directly enforce $\mathcal{L}_{\text{Rec}}$. Instead, we encourages the projected novel view Gaussians $\widehat{g}$ to be consistent with the synthesized novel view $\widehat{m}'$ via a second density loss: $\mathcal{L}_{\widehat{g}}(\widehat{m}', \widehat{g}) = \|\widehat{m}' - \sum_{k=1}^{K} \widehat{g}_k\|_1$. Without this loss, $g$ can describe well the input mask $m$, but the rotated $\widehat{g}$ may not describe well the generated mask $\widehat{m}'$.

**Self-supervised transform inverse 3D Gaussian loss.** We can also pass $m'$ back through our 3D Gaussian prediction stages (Fig. 2, green) to recover an estimate of the proxies under random transform $\mathbf{T}_R$. Then, we can invert this transform and penalize a loss against our initial estimate of the 3D Gaussians. With slight notation abuse: $\mathcal{L}_{\widehat{\mathcal{G}}}(\mathcal{G}, \widehat{\mathcal{G}}') = \|\mathcal{G} - \mathbf{T}_R^{-1}(\eta(\widehat{m}'))\|_1$, where $\eta$ predicts 3D Gaussians for a mask.

**Adversarial loss.** Training using only reconstruction losses tends to produce blurry images, so we adopt an adversarial training strategy. $G_m$ attempts to generate realistic masks to fool a discriminator $D_m$, while $D_m$ attempts to classify generated masks separately from real training masks. Within this, we also discriminate against our self-supervised transform masks $\widehat{m}'$: these should also fool $D_m$. We use a hinge-GAN loss $\mathcal{L}_{\text{Adv}}$ for better training stability [25, 50, 30]:

$$\mathcal{L}_{\text{Adv}}(G_m, D_m) = \mathbb{E}_{\widehat{m}'}[\min(0, -G_m(\widehat{m}') - 1)] + \quad (5)$$
$$2\mathbb{E}_m[\min(0, G_m(m) - 1)] + \mathbb{E}_{m'}[\min(0, -D_m(m') - 1)].$$

To reconstruct the 3D shape within a consistent world space, along with $m$ and $m'$, we find that it is sufficient to give the discriminator a mask $\widehat{m}'$ generated from only one random rotation per image (as similarly found by Henzler et al. [6]), rather than multiple random rotations.

**Feature match loss.** We improve sharpness by enforcing that real and generated images elicit similar deep feature

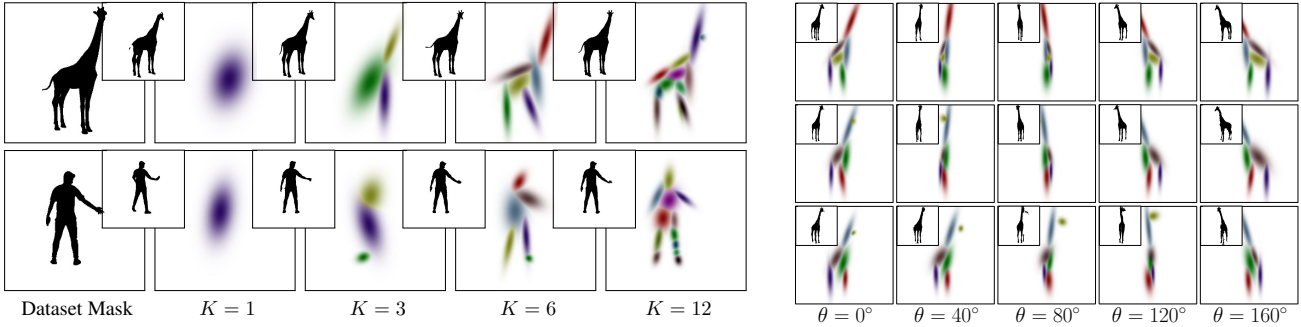

Figure 4. *Left:* Varying $K$ produces levels of abstraction over the object's shape and pose and so over generation control. At low $K$, only the major features are represented such as legs and neck. At higher $K$, details like individual legs ($K = 12$, top) and leg parts (calf, thigh) appear with the detail required to model the pose variation, e.g., in *Manuel* (bottom), the right leg moves more in the animation and gains a knee at $K = 12$. *Right:* Randomly sampling sparser datasets still recovers the coarse 3D structure of the input object. Rows 0, 1, 2, use $\frac{1}{16}, \frac{1}{32}, \frac{1}{64}$, of images in the training set; approximately 140, 70, and 35 images respectively. Colorings are different across rows.

responses in each layer $l$ of the discriminator $D_M^{(l)}$ [42, 56]:

$$\mathcal{L}_{\text{FM}}(D_{\boldsymbol{m}}) = \mathbb{E}_{\boldsymbol{m},\boldsymbol{m}',\widehat{\boldsymbol{m}}'} \left[ \Sigma_{l=1}^{L} \left\| D_{\boldsymbol{m}}^{(l)}(\widehat{\boldsymbol{m}}') - \bar{D}_{\boldsymbol{m}}^{(l)}(\boldsymbol{m}) \right\|_2^2 \right.$$
$$\left. + \left\| D_{\boldsymbol{m}}^{(l)}(\boldsymbol{m}') - \bar{D}_{\boldsymbol{m}}^{(l)}(\boldsymbol{m}) \right\|_2^2 \right], \quad (6)$$

where $\bar{D}_{\boldsymbol{m}}^{(l)}$ is the moving average of feature activations in layer $l$, and $L$ is the number of layers.

**Constraining pose and shape.** We bound $\boldsymbol{\mu}_k$ to $[-1, 1]$ and the diagonal values of $\boldsymbol{\Sigma}_k^c$ to $[0.01, 0.51]$. We prevent any Gaussian from being too small/too large; this encourages learning to use all Gaussians. To remove implausible canonical $\mathcal{G}^c$, we constrain $\mathbf{T}_k = (\mathbf{s}_k, \mathbf{t}_k, \boldsymbol{\theta}_k)$ to produce per-image $\mathcal{G}$ that remain somewhat close to $\mathcal{G}^c$ while still giving freedom to accommodate shape and pose changes (Figure 3). Discriminating masks generated from $\mathcal{G}^c$ is also possible, with self-supervision via random transforms, and may help relax per-image transform constraints.

### 3.2. Discussion

**Importance of losses and components (Figure 3).** Removing the reconstruction losses on $\boldsymbol{m}'$ allows a mask to only approximately correspond to the Gaussians as long as it satisfies the discriminator and $\mathcal{L}_{\boldsymbol{g}}$. Removing the density loss on $\boldsymbol{g}$ causes less 'coherent' Gaussians: they are not forced to represent the generated mask, yet $G_{\boldsymbol{m}}$ can still produces a high detail mask from these Gaussians. Finally, the transform inverse loss 'closes the loop' for the self supervision and helps maintain 3D space consistency and mask quality, especially under cases when penalizing the 2D maps $\widehat{g}'$ alone cannot accurately predict 3D, such as when objects have strong rotation-dependent self occlusion.

Canonical $\mathcal{G}^c$ encourages a meaningful 3D space as each image should be consistent with other images. Directly estimating per-image Gaussians fails for thin front/back views as the self-supervised rotation must only be consistent with $\mathcal{L}_{\boldsymbol{g}}$ and discrimination $\mathcal{L}_{\text{Adv,FM}}$ (Fig. 3e), instead learning a non-linear space that only rotates between front/back views.

Further, estimating $\mu, \Sigma$ values without the const+FC layers [13] led to worse performance.

Quantitatively, we compute IoU and DSSIM between ground truth and generated masks across a range of angles (Fig. 3, bottom). While removing $\mathcal{L}_{\widehat{\mathcal{G}}}$ or $\mathcal{G}^c$ improve the metrics slightly, qualitatively our final model is more coherent: parts can flicker in and out without $\mathcal{L}_{\widehat{\mathcal{G}}}$, and the 3D space is less coherent without $\mathcal{G}^c$.

**Varying $K$ and dataset size.** We provide $K$ at training time, which is simple to estimate by hand for many objects, e.g., one each for the body and head, one for each limb. As $K$ varies, our density losses over random rotations encourage detail where it is required (Fig. 4, left). Too few $K$ diminish pose or shape; too many $K$ leads to redundant Gaussians. As we set a minimum size, these appear as 'little dots' (Fig. 4, right) and can be ignored without affecting downstream tasks. For more control, a user could pre-define the canonical $\mathcal{G}^c$ from which a set of per-image deformations is learned. We also shows how the Gaussians are still usefully recovered as input data decreases $64\times$ in number (Fig. 4, right), though with less mask detail.

## 4. Mask texturing

We apply our 3D Gaussian proxies as a conditioning rig to an image generation task [47]. We demonstrate object posing and inserting into an existing image (Fig. 5). For this, we condition a second separately-trained GAN on the mask and on a background image to let us approximately match scene lighting.

Given a database of RGB images $\boldsymbol{i} \in \mathcal{I}$ and corresponding binary masks $\boldsymbol{m} \in \mathcal{M}$, we wish to learn a generative model of texture inside the mask conditioned on the background. First, we compute the background image $\boldsymbol{i}_b = \boldsymbol{i} \odot (\mathbf{1} - \boldsymbol{m})$ and the foreground image $\boldsymbol{i}_f = \boldsymbol{i} \odot \boldsymbol{m}$, where $\odot$ is the element-wise product. Next, we use an appearance encoder $E_{\boldsymbol{i}}$ to extract a latent representation $\mathbf{z}_{\boldsymbol{i}} \in \mathbb{R}^8$ for the foreground texture: $\mathbf{z}_{\boldsymbol{i}} = E_{\boldsymbol{i}}(\boldsymbol{i}_f)$; this lets us sample foregrounds at test time. We tile $\mathbf{z}_i$ and concatenate it with the background image $i_b$, and pass it into a U-Net-like network $G_i$ to generate texture. In $G_i$'s encoding phase, we layer-wise condition via $\mathbf{z}_i$. In $G_i$'s decoding phase, we concatenate the Gaussian maps $g$ obtained from $m$ and apply layer-wise conditioning as per $G_m$. The final image $i'$ is created from the output of $G_i$ with the original background: $i' = G_i(i_b, \mathbf{z}_i, g) \odot m + i_b$.

**Losses.** We train our network by minimizing an energy:

$$\mathcal{L}^i(G_i, E_i, D_{i,m}) = \beta_1 \mathcal{L}^i_{\text{Rec}} + \beta_2 \mathcal{L}^i_p + \beta_3 \mathcal{L}^i_{\text{KL}} + \beta_4 \mathcal{L}^i_{\text{Adv}} \\ + \beta_5 \mathcal{L}^i_{\text{FM}} + \beta_6 \mathcal{L}^i_{\text{zRec}}. \quad (7)$$

**Reconstruction loss.** We encourage the synthesized image $i'$ to be an identity of the input image $i$. We use the $L_1$ loss: $\mathcal{L}^i_{\text{Rec}}(i, i') = \|i - i'\|_1$.

**Perceptual loss.** We encourage finer detail by using a VGG16 perceptual loss [11] from the second convolutional block ($\phi_2 :=$'conv2$_2$'): $\mathcal{L}^i_p(i, i') = \|\phi_2(i) - \phi_2(i')\|_1$

**KL loss.** To structure $\mathbf{z}_i$ for test-time sampling, we predict mean and variance vectors for $\mathbf{z}_i$, sample one using the re-parametrization trick [17], then enforce that it comes from a Normal distribution using the KL divergence loss.

**Latent reconstruction loss.** The KL loss does not ensure that $G_i$ decodes $\mathbf{z}_i$ into diverse images. To prevent $\mathbf{z}_i$ from being ignored, we add a novel encoder $E'_i$ to reconstruct $\mathbf{z}_i$ from $i'$, and enforce a reconstruction loss via: $\mathcal{L}^i_{\text{zRec}} = \|\mathbf{z}_i - \mathbf{z}_{i'}\|_1$. When back-propagating gradients from $\mathcal{L}^i_{\text{zRec}}$, we update all generation parameters *apart* from those in $E_i$. This avoids $E_i$ and $G_i$ hiding the latent code information without producing diverse images [58].

**Adversarial losses.** Finally, we also train $D_{i,m}$ to discriminate $(i, m)$ from $(i', m)$ such that the RGB image and mask are correlated, and we use $D_{i,m}$ to penalize a feature matching loss between $(i, m)$ and $(i', m)$.

## 5. Experiments

**Datasets.** We render RGB images and masks using path tracing with ten real-world 360° HDR lighting maps of outdoor natural environments for realistic lighting and self-shadowing. For each image, we randomly rotate the camera around the up vector at a fixed distance from the object, to match settings in the literature [33]. We use four datasets without pose variation and of increasing shape complexity (*Maple*, *Airplane*, *Carla*, and *Pegasus*), and four animated datasets with pose variation (*Bee*, *Giraffe*, *Manuel*, *Old Robot*). These include hovering and flapping wings, walking, neck bending, and dancing (each with 110-400 frames; see video). We randomly sample animation frames: poses are not matched across views or in any temporal or rotation order, and we discard object and camera poses during training. We use 1,000/2,000 images for static/animated datasets, with a random 90/10% training/test split.

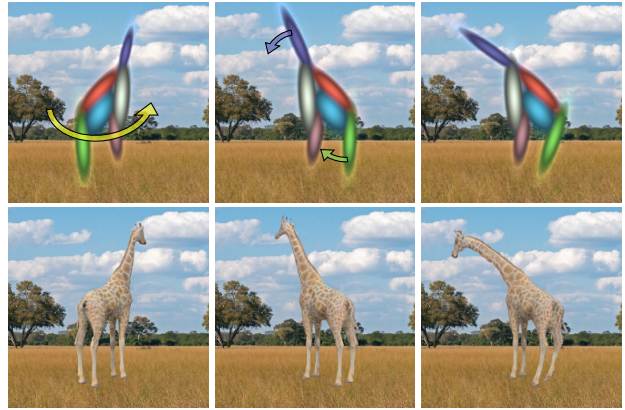

Figure 5. Explicitly recovering camera, shape, and pose allows interactive 3D Gaussian manipulation to generate novel instances.

**Training and hyperparameters.** We train mask and texture generators for 200 epochs on 2 RTX 2080 TI GPUs. We use the ADAM optimizer with a learning rate of $1e-4$, and $\beta = 0.5$. For static datasets, we predict the yaw rotation $R_\phi$ per image to affect the canonical Gaussians. For the mask hyper-parameters, we set $\lambda_1 = 100$, $\lambda_2 = 100$, $\lambda_3 = 100$, $\lambda_4 = 100$, $\lambda_5 = 1$, $\lambda_6 = 10$. We chose $\lambda_1, \lambda_2, \lambda_3$. $\lambda_4$ over the interval $[0, 10, 50, 100]$. For *Giraffe* with slower animation, $\lambda_3 = 10$ and $\lambda_4 = 50$ led to a slightly better Gaussians. For texture hyper-parameters, we fix $\beta_1 = 100$ $\beta_2 = 0.5$ $\beta_3 = 0.01$ $\beta_4 = 1$ $\beta_5 = 10$ $\beta_6 = 0.1$.

**Baselines.** To show the value of model components, we compare to HoloGAN [33], PlatonicGAN [6], and Liao et al. [24], and provide methods with *just masks* and *just RGB foregrounds*. Each uses 3D proxies to generate images. HoloGAN and PlatonicGAN use voxels: the HoloGAN bottleneck has 64-dim. deep appearance vectors in $16^3$ voxels that are projected to 2D and decoded, while PlatonicGAN directly predicts a $64^3$ RGBA voxel space *per image* to handle variation. As a constraint, HoloGAN employs a weaker (but flexible) *latent* reconstruction loss vs. our projection and pixel-wise reconstruction loss. PlatonicGAN also uses a pixel-wise reconstruction loss and, via projection, this can be related to unposed voxel carving when given masks as inputs. Liao et al. estimate multiple 3D primitives (cubes or spheres) as proxies for objects with simple geometry. Due to limited space, please see our supplemental material for Liao et al. results and comparisons to unsupervised 2D part maps of Lorenz et al. [28].

**Results—mask only.** This setting compares the ability to reconstruct a 3D camera and object space. For the static datasets (Fig. 7), HoloGAN's deep voxels reconstruct the input masks, but its latent rotation space can be incoherent with masks at incorrect angles. PlatonicGAN's voxel spaces are naturally 3D and with shape detail, but suffer some incorrect rotations and include spurious or missing geometry. Our approach infers plausible coarse 3D structure that con-trols high-quality 2D mask generation.

| On masks | IoU×100 ▲ | DSSIM×100 ▼ |
|---|---|---|
| Ours | **81.97** | **9.35** |
| PlatonicGAN [6] | 77.29 | 21.29 |
| On RGB | KID×100 ▼ | FID×100 ▼ |
| Ours (via masks) | **9.16 ± 0.60** | **117.81** |
| PlatonicGAN [6] | 49.7 ± 0.89 | 375.26 |
| HoloGAN [33] | 32.72 ± 0.87 | 298.35 |
| Liao et al. [24] | 34.2 ± 0.84 | 292.89 |

Table 1. Metrics are computed per dataset and then mean averaged (see supplemental for full results).

For the animated datasets, methods must accommodate image pose variation, and all baselines perform worse. HoloGAN has both part errors (incorrect leg placement) and a low-coherence 3D space (rotation is not smooth; Fig. 1). PlatonicGAN's shapes are incorrectly reconstructed, with missing or misplaced legs and spurious content: even though the method estimates per-image shape, without a canonical model these are incorrectly corresponded in the 3D estimation task, combining parts of objects from across poses. Our method by construction has a canonical 3D $\mathcal{G}^c$ and transformable parts, producing a coherent 3D camera and coarse posable object space.

**Results—foreground only.** Here, the reconstruction task is more complex, with additional texture and lighting variation. Even for static scenes, HoloGAN struggles to generate high-quality appearance, and the resulting 3D spaces for dynamic scenes mix all input variations (Fig. 1) or fail to correctly rotate the image (Fig. 7, *Maple*). PlatonicGAN successfully generates detail (Fig. 1) but again these have object geometry errors and the predicted voxel coloring only approximates the intended output (Fig. 7, *all*). Liao et al. generated images are of broadly good quality, though the pose is entangled with the camera rotation and texture is low resolution and less consistent. As might be expected, our approach demonstrates that using additional mask information to separate shape and appearance allows conditioning higher-fidelity 2D texture generation with disentangled 3D camera, pose, and lighting consistency.

**Quantitative results.** PlatonicGAN and our method infer explicit 3D spaces. As such, we compute IoU and DSSIM on masks at a known camera angle and compare to test-set ground truth masks: if a method forms a coherent 3D camera and object space, then masks will match (Tab. 1). For methods that infer implicit 3D spaces (without meaningful angles), we compute KID and FID on generated RGB foregrounds (KID/FID are pre-trained for RGB via ImageNet).

**Real-world data.** We show the benefits of a mid-level 3D structure via the managed control of variation available in synthetic data. Many other variations exist in real-world datasets. To show this gap, we demonstrate our method on highly-varied MS COCO data (Fig. 6). Here, our 3D space and Gaussians are plausible, even though there is significant quality variation in the hand-drawn input masks especially for front/back views; better masks would improve this [18].

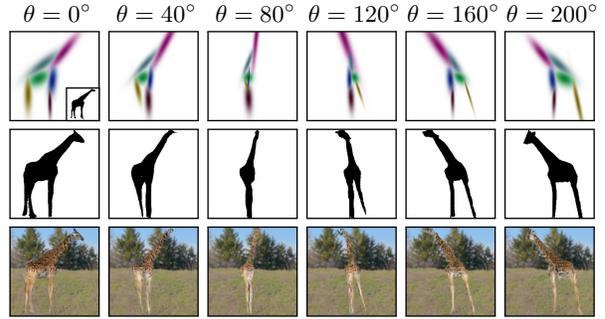

Figure 6. Our method can produce plausible 3D Gaussians from low-quality and highly-varied masks from MS COCO.

## 6. Discussion and Limitations

**3D/2D Bridge.** Our approach infers a coarse 3D shape and pose as a 'rig' for 2D mask and texture generators, rather than an explicit representation of an object's detailed 3D appearance. We show that making this trade-off can lead to more coherent 3D spaces. Our intermediate structure is simple, robust, and allows interactive control over camera and pose for image generation via 2D masks and textures.

**Objectness.** This approach can apply to many objects as it assumes no explicit connectivity like a skeleton. Adding hierarchies [10] or kinematic chains could constrain the per-Gaussian transforms, though this is challenging without a template. Further, a Gaussian transmittance function [41] could help resolve front/back ambiguity for better texture.

**Intra-class variation.** While our method can handle intra-class pose and some appearance variation (lighting), future work should explore additional structure to handle variation of objects within the same class but that vary in their parts, or that have underlying albedo variation.

**Illusions.** Finally, humans can be fooled by silhouette illusions like the Spinning Dancer that appears to rotate either way. For certain objects and a sufficiently-small $K$, our output 3D space can also end up 'rotated' even though it is smooth, i.e., an input mask representing $45°$ is reconstructed at $315°$. This can be fixed by reflecting the space.

## 7. Conclusion

As we move toward 'in the wild' settings, we need intermediate structures for varied objects and training losses that can produce meaningful 3D spaces. Given unposed silhouettes, we reconstruct a coarse Gaussian-based representation of 3D camera and object space along with variations in object pose. We discuss trade-offs with two voxel-based baselines, and display a potential use of our approach to condition 2D texture generators. Future work could explore deep appearance Gaussians and additional ways to self supervise reconstruction under variation to move closer still to handling complex objects in natural images.

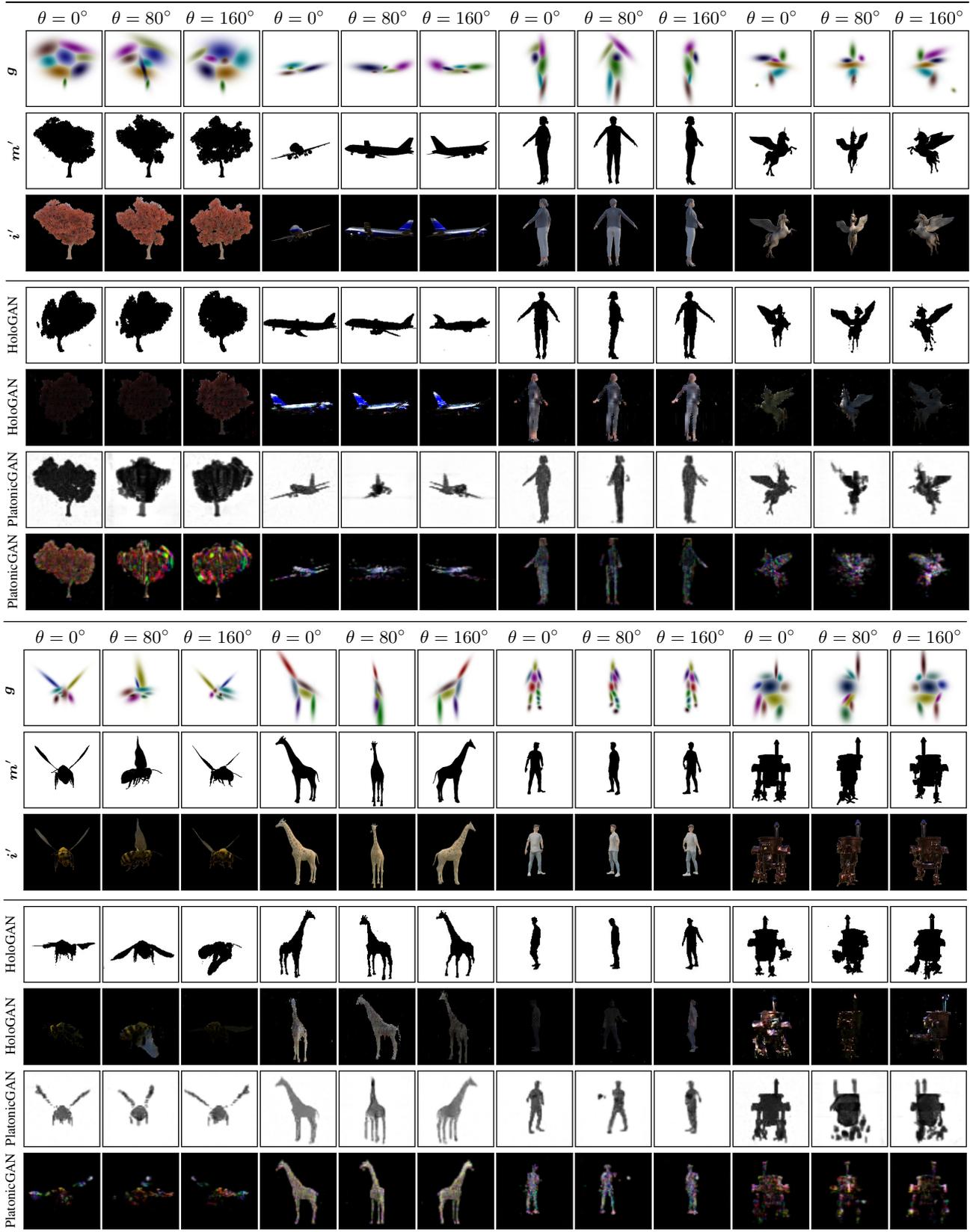

Figure 7. Please zoom in to see detail. *Rows in each block:* Reconstructed Gaussians, masks, and RGB images, across three output angles and with any texture-specific latent variables fixed, with comparisons to HoloGAN [33] and PlatonicGAN [6] run on *just masks* and *just RGB foregrounds*. Note that HoloGAN only infers a latent 'angle', making mapping to an explicit 3D space not possible. *Top block of five rows:* Datasets of objects of fixed pose showing increasing shape complexity: *Maple, Airplane, Carla, Pegasus*. *Bottom block of five rows:* Datasets of animated objects with varying pose showing increasing shape complexity: *Bee, Giraffe, Manuel, Old Robot*.

**Acknowledgements** Thank you to Numair Khan for creating the Maya synthetic dataset generator, and for fruitful discussions with Helge Rhodin, Srinath Sridhar, Michael Snower, Yuanhao Wang, Stephen H. Bach, and Daniel Ritchie. Thanks also to Insoo Kim and Neil Sehgal. KIK was supported by the National Research Foundation of Korea (NRF) grant (No. 2021R1A2C2012195). YAM thanks the Marie Sklodowska-Curie grant No 665992, and the Centre for Doctoral Training in Digital Entertainment (CDE), EP/L016540/1. This material is based on research sponsored by Defense Advanced Research Projects Agency (DARPA) and Air Force Research Laboratory (AFRL) under agreement number FA8750-19-2-1006. The U.S. Government is authorized to reproduce and distribute reprints for Governmental purposes notwithstanding any copyright notation thereon. The views and conclusions contained herein are those of the authors and should not be interpreted as necessarily representing the official policies or endorsements, either expressed or implied, of Defense Advanced Research Projects Agency (DARPA) and Air Force Research Laboratory (AFRL) or the U.S. Government.

## References

[1] Panos Achlioptas, Olga Diamanti, Ioannis Mitliagkas, and Leonidas Guibas. Learning representations and generative models for 3D point clouds. In *ICML*, pages 40–49. PMLR, 2018. 2

[2] Kunihiro Baba, Ritei Shibata, and Masaaki Sibuya. Partial correlation and conditional correlation as measures of conditional independence. *Australian & New Zealand Journal of Statistics*, 46(4):657–664, 2004. 12, 13, 15

[3] Boyang Deng, Kyle Genova, Soroosh Yazdani, Sofien Bouaziz, Geoffrey Hinton, and Andrea Tagliasacchi. Cvxnet: Learnable convex decomposition. In *Proceedings of the IEEE/CVF Conference on Computer Vision and Pattern Recognition (CVPR)*, June 2020. 2

[4] Kyle Genova, Forrester Cole, Avneesh Sud, Aaron Sarna, and Thomas Funkhouser. Local deep implicit functions for 3D shape. In *CVPR*, pages 4857–4866, 2020. 2

[5] Ian J. Goodfellow, Jean Pouget-Abadie, Mehdi Mirza, Bing Xu, David Warde-Farley, Sherjil Ozair, Aaron Courville, and Yoshua Bengio. Generative adversarial nets. In *NeurIPS*, 2014. 2

[6] Philipp Henzler, Niloy J Mitra, and Tobias Ritschel. Escaping Plato's cave: 3D shape from adversarial rendering. In *ICCV*, pages 9984–9993, 2019. 1, 2, 3, 4, 6, 7, 8, 15

[7] Seunghoon Hong, Xinchen Yan, Thomas Huang, and Honglak Lee. Learning hierarchical semantic image manipulation through structured representations. In *NeurIPS*, 2018. 2

[8] Tomas Jakab, Ankush Gupta, Hakan Bilen, and Andrea Vedaldi. Unsupervised learning of object landmarks through conditional image generation. In *NeurIPS*, 2018. 2

[9] Tomas Jakab, Ankush Gupta, Hakan Bilen, and Andrea Vedaldi. Self-supervised learning of interpretable keypoints from unlabelled videos. In *CVPR*, 2020. 2

[10] Farnoosh Javadi Fishani. Hierarchical part-based disentanglement of pose and appearance. Master's thesis, University of British Columbia, 2020. 7

[11] Justin Johnson, Alexandre Alahi, and Li Fei-Fei. Perceptual losses for real-time style transfer and super-resolution. In *ECCV*, 2016. 6, 17

[12] Angjoo Kanazawa, Shubham Tulsiani, Alexei A Efros, and Jitendra Malik. Learning category-specific mesh reconstruction from image collections. In *ECCV*, pages 371–386, 2018. 2

[13] Tero Karras, Samuli Laine, and Timo Aila. A style-based generator architecture for generative adversarial networks. *CVPR*, 2019. 3, 5

[14] Tero Karras, Samuli Laine, Miika Aittala, Janne Hellsten, Jaakko Lehtinen, and Timo Aila. Analyzing and improving the image quality of stylegan. In *CVPR*, pages 8110–8119, 2020. 2

[15] Hiroharu Kato and Tatsuya Harada. Self-supervised learning of 3d objects from natural images. *arXiv preprint arXiv:1911.08850*, 2019. 2

[16] Yunji Kim, Seonghyeon Nam, In Cho, and Seon Joo Kim. Unsupervised keypoint learning for guiding class-conditional video prediction. In *NeurIPS*, 2019. 2

[17] Diederik P Kingma and Max Welling. Auto-encoding variational bayes. *ICLR*, 2014. 6, 17

[18] Alexander Kirillov, Yuxin Wu, Kaiming He, and Ross Girshick. Pointrend: Image segmentation as rendering. In *Proceedings of the IEEE/CVF conference on computer vision and pattern recognition*, pages 9799–9808, 2020. 7

[19] Markus Knoche, István Sárándi, and Bastian Leibe. Reposing humans by warping 3D features. In *CVPR Workshop on Towards Human-Centric Image/Video Synthesis*, 2020. 2

[20] Nilesh Kulkarni, Abhinav Gupta, David F Fouhey, and Shubham Tulsiani. Articulation-aware canonical surface mapping. In *CVPR*, pages 452–461, 2020. 2

[21] Donghoon Lee, Sifei Liu, Jinwei Gu, Ming-Yu Liu, Ming-Hsuan Yang, and Jan Kautz. Context-aware synthesis and placement of object instances. In *NIPS*, pages 10393–10403, 2018. 2

[22] Jiahui Lei, Srinath Sridhar, Paul Guerrero, Minhyuk Sung, Niloy Mitra, and Leonidas J. Guibas. Pix2Surf: Learning parametric 3D surface models of objects from images. In *ECCV*, 2020. 2

[23] Yuheng Li, Krishna Kumar Singh, Utkarsh Ojha, and Yong Jae Lee. MixNMatch: Multifactor disentanglement and encoding for conditional image generation. In *CVPR*, 2020. 2

[24] Yiyi Liao, Katja Schwarz, Lars Mescheder, and Andreas Geiger. Towards unsupervised learning of generative models for 3D controllable image synthesis. In *CVPR*, 2020. 2, 6, 7, 12, 13, 15

[25] Jae Hyun Lim and Jong Chul Ye. Geometric GAN. *arXiv preprint arXiv:1705.02894*, 2017. 4

[26] Chen-Hsuan Lin, Ersin Yumer, Oliver Wang, Eli Shechtman, and Simon Lucey. ST-GAN: Spatial transformer generative adversarial networks for image compositing. In *CVPR*, 2018. 2

[27] Tony Lindeberg. Image matching using generalized scale-space interest points. *Journal of Mathematical Imaging and Vision*, 52(1):3–36, 2015. 2

[28] Dominik Lorenz, Leonard Bereska, Timo Milbich, and Bjorn Ommer. Unsupervised part-based disentangling of object shape and appearance. In *CVPR*, pages 10955–10964, 2019. 2, 6, 12, 14

[29] Ben Mildenhall, Pratul P. Srinivasan, Matthew Tancik, Jonathan T. Barron, Ravi Ramamoorthi, and Ren Ng. Nerf: Representing scenes as neural radiance fields for view synthesis. In *ECCV*, 2020. 2

[30] Takeru Miyato, Toshiki Kataoka, Masanori Koyama, and Yuichi Yoshida. Spectral normalization for generative adversarial networks. In *ICLR*, 2018. 4

[31] Natalia Neverova, Artsiom Sanakoyeu, Patrick Labatut, David Novotny, and Andrea Vedaldi. Discovering relationships between object categories via universal canonical maps. In *CVPR*, 2021. 2

[32] Alejandro Newell, Kaiyu Yang, and Jia Deng. Stacked hourglass networks for human pose estimation. In *ECCV*, 2016. 2

[33] Thu Nguyen-Phuoc, Chuan Li, Lucas Theis, Christian Richardt, and Yong-Liang Yang. HoloGAN: Unsupervised learning of 3D representations from natural images. In *ICCV*, pages 7588–7597, 2019. 1, 2, 6, 7, 8, 15

[34] Michael Niemeyer and Andreas Geiger. Giraffe: Representing scenes as compositional generative neural feature fields. In *Proceedings of the IEEE/CVF Conference on Computer Vision and Pattern Recognition*, pages 11453–11464, 2021. 2

[35] Chengjie Niu, Jun Li, and Kai Xu. Im2struct: Recovering 3d shape structure from a single rgb image. In *Proceedings of the IEEE conference on computer vision and pattern recognition*, pages 4521–4529, 2018. 2

[36] Kyle Olszewski, Sergey Tulyakov, Oliver Woodford, Hao Li, and Linjie Luo. Transformable bottleneck networks. In *Proceedings of the IEEE/CVF International Conference on Computer Vision*, pages 7648–7657, 2019. 2

[37] Pavel Ostyakov, Roman Suvorov, Elizaveta Logacheva, Oleg Khomenko, and Sergey I. Nikolenko. SEIGAN: Towards compositional image generation by simultaneously learning to segment, enhance, and inpaint. *arXiv preprint arXiv:1811.07630*, 2018. 2

[38] Taesung Park, Ming-Yu Liu, Ting-Chun Wang, and Jun-Yan Zhu. Semantic image synthesis with spatially-adaptive normalization. In *CVPR*, 2019. 2

[39] Varun Ramakrishna, Daniel Munoz, Martial Hebert, James Andrew Bagnell, and Yaser Sheikh. Pose machines: Articulated pose estimation via inference machines. In *ECCV*, 2014. 2

[40] Jian Ren, Menglei Chai, Oliver J Woodford, Kyle Olszewski, and Sergey Tulyakov. Flow guided transformable bottleneck networks for motion retargeting. In *Proceedings of the IEEE/CVF Conference on Computer Vision and Pattern Recognition*, pages 10795–10805, 2021. 2

[41] Helge Rhodin, Nadia Robertini, Christian Richardt, Hans-Peter Seidel, and Christian Theobalt. A versatile scene model with differentiable visibility applied to generative pose estimation. In *ICCV*, pages 765–773, 2015. 7

[42] Tim Saliman, Ian Goodfellow, Wojciech Zaremba, and Vicki Cheung. Improved techniques for training GANs. In *NeurIPS*, 2016. 5

[43] Katja Schwarz, Yiyi Liao, Michael Niemeyer, and Andreas Geiger. GRAF: Generative radiance fields for 3D-aware image synthesis. In *NeurIPS*, 2020. 2

[44] Aliaksandr Siarohin, Stéphane Lathuilière, Sergey Tulyakov, Elisa Ricci, and Nicu Sebe. Animating arbitrary objects via deep motion transfer. In *The IEEE Conference on Computer Vision and Pattern Recognition (CVPR)*, June 2019. 2

[45] Vincent Sitzmann, Justus Thies, Felix Heide, Matthias Nießner, Gordon Wetzstein, and Michael Zollhofer. Deepvoxels: Learning persistent 3D feature embeddings. In *CVPR*, pages 2437–2446, 2019. 2

[46] Srinath Sridhar, Helge Rhodin, Hans-Peter Seidel, Antti Oulasvirta, and Christian Theobalt. Real-time hand tracking using a sum of anisotropic gaussians model. In *3DV*, pages 319–326, 2014. 3, 27

[47] Ayush Tewari, Mohamed Elgharib, Gaurav Bharaj, Florian Bernard, Hans-Peter Seidel, Patrick Pérez, Michael Zollhofer, and Christian Theobalt. Stylerig: Rigging StyleGAN for 3D control over portrait images. In *CVPR*, pages 6142–6151, 2020. 5

[48] James Thewlis, Hakan Bilen, and Andrea Vedaldi. Unsupervised learning of object frames by dense equivariant image labelling. In *NIPS*, 2017. 2

[49] James Thewlis, Hakan Bilen, and Andrea Vedaldi. Unsupervised learning of object landmarks by factorized spatial embeddings. In *ICCV*, 2017. 2

[50] Dustin Tran, Rajesh Ranganath, and David M. Blei. Deep and hierarchical implicit models. *arXiv preprint arXiv:1702.08896*, 7, 2017. 4

[51] Yi-Hsuan Tsai, Xiaohui Shen, Zhe Lin, Kalyan Sunkavalli, Xin Lu, and Ming-Hsuan Yang. Deep image harmonization. In *CVPR*, 2017. 2

[52] Shubham Tulsiani, Hao Su, Leonidas J. Guibas, Alexei A. Efros, and Jitendra Malik. Learning shape abstractions by assembling volumetric primitives. In *Computer Vision and Pattern Regognition (CVPR)*, 2017. 2

[53] Mehmet Ozgur Turkoglu, William Thong, Luuk Spreeuwers, and Berkay Kicanaoglu. A layer-based sequential framework for scene generation with GANs. In *AAAI*, pages 8901–8908, 2019. 2

[54] Kohei Yamashita, Shohei Nobuhara, and Ko Nishino. 3D-GMNet: Single-view 3D shape recovery as a gaussian mixture. *BMVC*, 2020. 2, 3

[55] Jianwei Yang, Anitha Kannan, Dhruv Batra, and Devi Parikh. LR-GAN: Layered recursive generative adversarial networks for image generation. In *ICLR*, 2017. 2

[56] Richard Zhang, Phillip Isola, Alexei A. Efros, Eli Shechtman, and Oliver Wang. The unreasonable effectiveness of

deep features as a perceptual metric. In *CVPR*, pages 586–595, 2018. 5

[57] Jun-Yan Zhu, Taesung Park, Phillip Isola, and Alexei A. Efros. Unpaired image-to-image translation using cycle-consistent adversarial networks. In *ICCV*, 2017. 2

[58] Jun-Yan Zhu, Richard Zhang, Deepak Pathak, Trevor Darrell, Alexei A. Efros, Oliver Wang, and Eli Shechtman. Toward multimodal image-to-image translation. In *NeurIPS*, 2017. 6, 17

[59] Silvia Zuffi, Angjoo Kanazawa, Tanya Berger-Wolf, and Michael J. Black. Three-D safari: Learning to estimate zebra pose, shape, and texture from images "in the wild". In *ICCV*, pages 5358–5367, 2019. 2

# Supplemental Material

This supplemental document contains additional comparisons against an unsupervised 2D parts-based inference method [28] and on a geometry proxy-based method for scene generation [24] that could not fit into the main paper (Sec. A). We also provide additional and fuller results reporting (Sec. B), more detailed discussion of how to estimate Gaussian covariance matrices (Sec. C), additional details on mask texturing (Sec. D), network architecture details (Sec. E), and finally a derivation of the analytic Gaussian projection derivative (Sec. F).

Please also see our supplemental video, which includes examples of interactive editing and results showing rotations of the recovered 3D Gaussians and their use in generating masks and textures, in comparison to HoloGAN, PlatonicGAN, and Liao et al.

## A. Additional comparisons

### A.1. Liao et al.'s [24] method on our data

Liao et al. [24]'s method uses cube and sphere mesh proxies to represent multiple simple scene objects, and this allows control in image generation over camera rotation and object rotation. Our data has one complex object with deforming parts. Figure 8 shows that these proxies are promising but still unfortunately too simple for our data. Even though their generated images are of reasonable quality (though low in resolution and with minor artifacts), the pose of the object changes when rotating, *e.g.* the *Giraffe* neck bends while rotating. One dataset fails to rotate at all—*Pegasus*. Further, while their RGB appearance is often as vivid as the input, the texture is not consistent when rotating the camera, e.g., lighting variation, as the proxy geometry is not sufficiently descriptive to allow separation of shape and texture variation. Liao et al. is designed primarily for multiple objects; future work could build upon both methods to handle multiple complex objects.

### A.2. Lorenz et al. [28] 2D part discovery on our data

For parts-based discovery, we compare our 3D parts to the 2D part maps from Lorenz et al. [28]. For fair comparison, we train the method only on mask images. The discovered parts are relevant, though some areas miss representation (Fig. A.2, *Maple*), and with less conformance to the underlying 3D space (e.g., failing to rotate correctly with the object, *Carla*).

## B. Additional results

We show additional results for Giraffe, Manuel, Bee, Old Robot, Carla, Maple, Airplane and Pegasus in Figs. 13–20.

**Interactive editing.** In our supplemental video, we show an application of user control over the recovered Gaussian proxies. We built a demo that exploits our Gaussian proxy's form to allow simple drag-and-drop object part translation, anisotropic scale, and rotation, plus camera control and lighting variation via $z_i$. This allows in-distribution editing of the poses, e.g., *Giraffe* neck bending and leg adjustments. It also allows some out-of-distribution adjustments, such as placing the giraffe in a combination of poses that were not in any one input example, or more 'creative' edits such as enlarging or elongating certain Gaussians.

For our interactive editing scenario, which is 2D, each individual generated 2D image is coherent. However, while our approach recovers a coarse 3D 'rig' or 'artist's mannequin' for an object, some texturing can be inconsistent in 3D (such as in rotation animations) as we only affect a 2D generator. While this gives high resolution, some fine detail can appear to shimmer as the camera rotates (e.g., on the *Maple* scene). This is because the chosen $K$ coarse Gaussians do not provide sufficiently localized conditioning for the small leaf features as the camera varies by small angles.

**Adaptation to background changes.** Figure 11 shows that when the background becomes darker, the generated foreground also becomes darker, and vice versa. In addition, we show in Figure 12 that our generated texture is still reasonable when given backgrounds that are out of distribution. This harmonization cannot be achieved by independently synthesizing foreground objects and placing them on a target background.

**Quantitative evaluations.** In Tables 2 and 3, we report quantitative metrics for each dataset independently; these were averaged in the main paper due to space limitations.

## C. Methods to estimate Gaussian covariance $\Sigma$

When estimating the covariance matrices $\Sigma$ of our Gaussian proxies, we consider three approaches: the eigendecomposition approach, the Cholesky decomposition approach, and the conditional covariance method [2].

**Eigendecomposition approach.** Naïvely predicting the values in the Gaussian covariance matrices $\Sigma_k$ as free parameters does not satisfy the positive definiteness requirements for a covariance matrix. Instead, we leverage the eigendecomposition of $\Sigma = \mathbf{V}\mathbf{U}\mathbf{V}^\top$, where $\mathbf{U}$ is the diagonal matrix of eigenvalues with strictly positive values on the diagonal, and $\mathbf{V}$ is an orthogonal matrix formed by the eigenvectors of $\Sigma$. We use a fully connected network to predict the diagonal values in $\mathbf{U}$. To ensure that they are positive, we use a sigmoid activation at the final layer, and also add a small $\epsilon = 0.01$ for strict positiveness. Similarly, we predict the columns of $\mathbf{V}$ using a fully connected network. In this case, we want $\mathbf{V}$ to be orthonormal. As such, we adopt the following process: First, we predict two vectors $\mathbf{v}_1$ and $\mathbf{v}_2'$ and obtain $\mathbf{v}_2$ as the cross product of $\mathbf{v}_1$ and $\mathbf{v}'_2$. Then, the third vector $\mathbf{v}_3$ is obtained as the cross product of $\mathbf{v}_1$ and $\mathbf{v}_2$. Finally, the $i$-th column of $\mathbf{V}$ is obtained by normalizing $\mathbf{v}_i$.

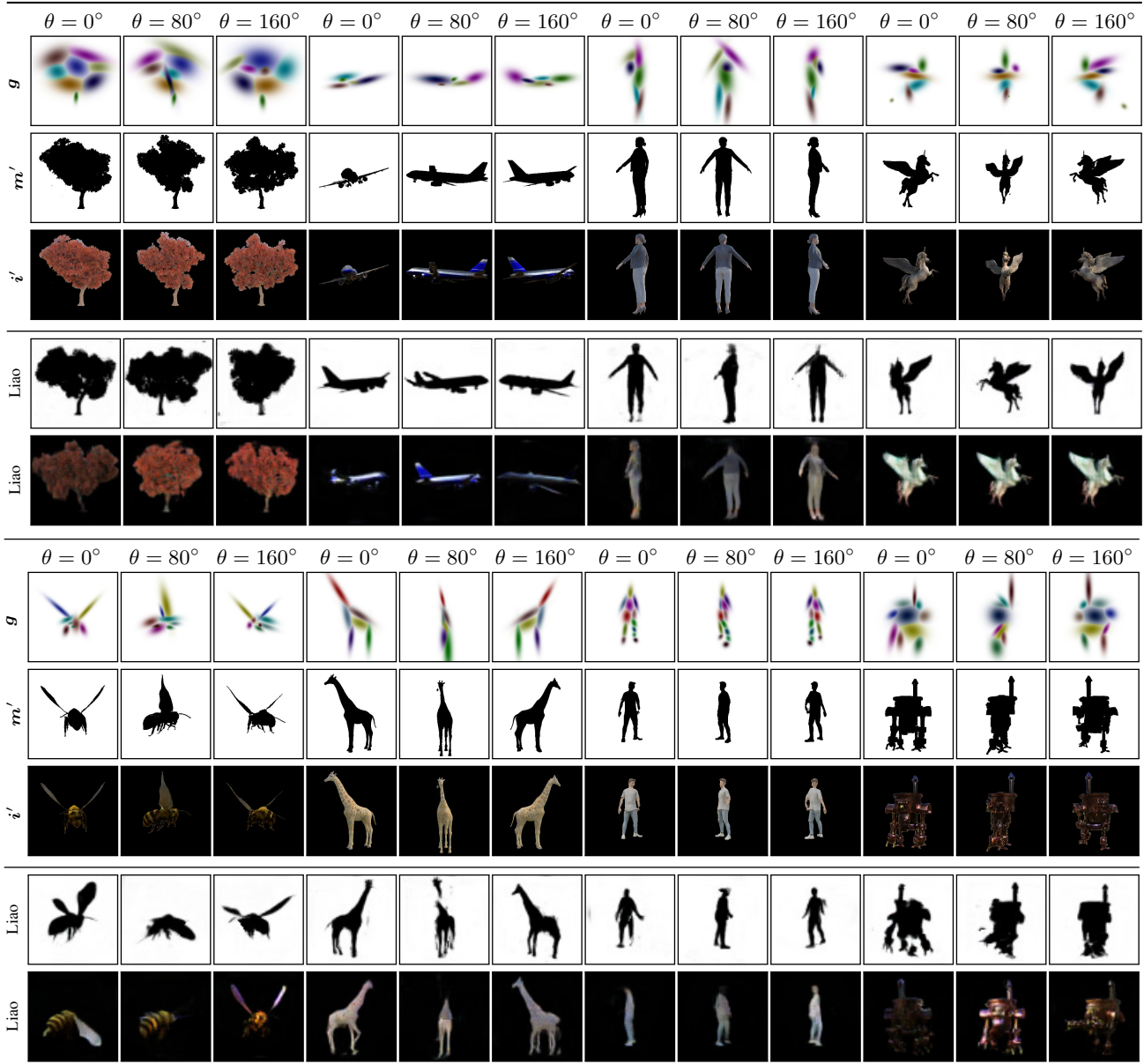

Figure 8. Please zoom in to see detail. *Rows in each block:* Reconstructed Gaussians, masks, and RGB images, across three output angles and with any texture-specific latent variables fixed, with comparisons to Liao *et al.* [24] and on *just masks* and *just RGB foregrounds*. *Top block of five rows:* Datasets of objects of fixed pose showing increasing shape complexity: *Maple, Airplane, Carla, Pegasus*. *Bottom block of five rows:* Datasets of animated objects with varying pose showing increasing shape complexity: *Bee, Giraffe, Manuel, Old Robot*.

In addition, learning covariances with 32-bit float types caused issues; 64-bit double produced more stable training.

**Cholesky decomposition approach.** The Cholesky decomposition approach enforces positive semi-definiteness via predicting $\mathbf{\Sigma}^{\frac{1}{2}}$ such that $\mathbf{\Sigma} = (\mathbf{\Sigma}^{\frac{1}{2}})^\top \mathbf{\Sigma}^{\frac{1}{2}}$. However, in our Gaussian proxies, the individual elements $\sigma$ of $\mathbf{\Sigma}$ have intended meaning as the 3D scale of object parts, but $\mathbf{\Sigma}^{\frac{1}{2}}$ does not provide an intuitive control over those $\sigma$ values.

**Conditional covariance approach.** In this approach [2], we describe the covariance matrix as:

$$\mathbf{\Sigma_i} = \begin{bmatrix} \sigma_1^2 & c_{12}\sigma_1\sigma_2 & c_{13}\sigma_1\sigma_3 \\ c_{12}\sigma_1\sigma_2 & \sigma_2^2 & c_{23}\sigma_2\sigma_3 \\ c_{13}\sigma_1\sigma_3 & c_{23}\sigma_2\sigma_3 & \sigma_3^2 \end{bmatrix}, \qquad (8)$$

where $\sigma_i$ are the individual standard deviations and $c_{ij}$ are the correlations between variables indexed by $i$ and $j$.

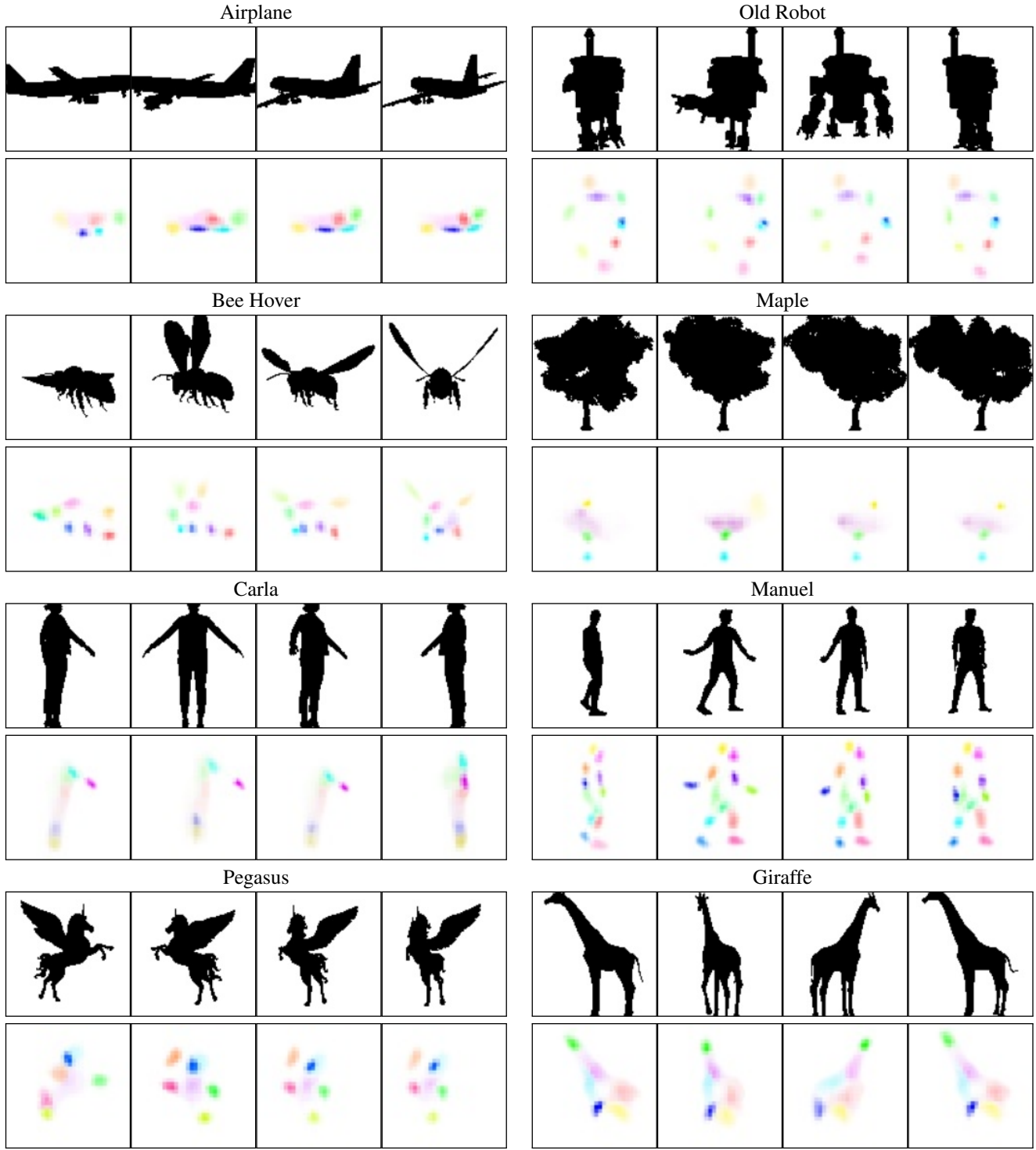

Figure 9. Results for *Lorenz et al.* [28] trained on masks only from our various datasets. Each has been trained for 50,000 iterations. The top row in each subfigure is the input into the network and the bottom row is the learned 2D part map. The part maps generally correspond to different areas of the object, but can struggle to represent object rotation (e.g., legs do not rotate in Manuel). For the 'dense' class of *Maple*, only the the relatively static trunk is consistently labeled.

Table 2. *Top:* KID × 100 ± STD × 100 and FID values (lower is better). *Bottom:* Mean IoU × 100 (higher is better) and DSSIM × 100 (lower is better) with respect to the reconstructed masks over the test set.

| On RGB | Giraffe | | Manuel | | Maple | | Carla | |
|---|---|---|---|---|---|---|---|---|
| | KID ▼ | FID ▼ | KID ▼ | FID ▼ | KID ▼ | FID ▼ | KID ▼ | FID ▼ |
| Ours | 2.72 ± 0.41 | 53.96 | 4.24 ± 0.35 | 62.36 | 11.39 ± 0.8 | 108.07 | 3.64 ± 0.29 | 62.73 |
| PlatonicGAN [6] | 42.98 ± 0.68 | 327.99 | 63.49 ± 0.77 | 435.38 | 53.95 ± 1.29 | 364.83 | 43.48 ± 0.64 | 338.58 |
| HoloGAN [33] | 39.28 ± 0.72 | 320.65 | 31.21 ± 0.81 | 292.44 | 46.49 ± 1.14 | 324.37 | 25.05 ± 0.68 | 240.78 |
| Liao et al. [24] | 33.43 ± 0.77 | 281.26 | 42.61 ± 0.86 | 317.13 | 36.56 ± 1.08 | 248.64 | 39.13 ± 0.81 | 310.21 |
| On masks | IoU ▲ | DSSIM ▼ | IoU ▲ | DSSIM ▼ | IoU ▲ | DSSIM ▼ | IoU ▲ | DSSIM ▼ |
| Ours | 83.96 | 6.22 | 81.46 | 5.66 | 86.51 | 21.87 | 89.39 | 6.22 |
| PlatonicGAN [6] | 67.70 | 23.85 | 67.54 | 16.57 | 92.77 | 33.09 | 83.66 | 13.20 |
| HoloGAN [33] | 30.68 | 39.38 | 42.62 | 18.40 | 62.74 | 64.24 | 38.60 | 38.83 |
| Liao et al. [24] | 31.27 | 60.82 | 41.13 | 31.14 | 68.94 | 88.61 | 36.50 | 58.93 |

Table 3. *Top:* KID × 100 ± STD × 100 and FID values (lower is better). *Bottom:* Mean IoU × 100 (higher is better) and DSSIM × 100 (lower is better) with respect to the reconstructed masks over the test set.

| On RGB | Bee | | Pegasus | | Old Robot | | Airplane | |
|---|---|---|---|---|---|---|---|---|
| | KID ▼ | FID ▼ | KID ▼ | FID ▼ | KID ▼ | FID ▼ | KID ▼ | FID ▼ |
| Ours | 7.62 ± 0.44 | 124.18 | 11.03 ± 0.61 | 156.33 | 15.93 ± 0.76 | 186.81 | 16.71 ± 1.17 | 188.07 |
| PlatonicGAN [6] | 38.06 ± 0.73 | 324.56 | 59.86 ± 1.29 | 456.46 | 56.49 ± 0.94 | 439.7 | 39.31 ± 0.79 | 314.61 |
| HoloGAN [33] | 18.32 ± 0.44 | 252.95 | 27.94 ± 0.94 | 291.92 | 41.29 ± 1.34 | 372.52 | 32.18 ± 0.91 | 291.29 |
| Liao et al. [24] | 24.77 ± 0.59 | 265.52 | 32.61 ± 0.76 | 325.35 | 37.83 ± 1.04 | 333.02 | 26.64 ± 0.82 | 261.98 |
| On masks | IoU ▲ | DSSIM ▼ | IoU ▲ | DSSIM ▼ | IoU ▲ | DSSIM ▼ | IoU ▲ | DSSIM ▼ |
| Ours | 76.11 | 9.22 | 85.17 | 6.01 | 87.58 | 9.86 | 65.57 | 9.72 |
| PlatonicGAN [6] | 65.82 | 23.76 | 82.35 | 17.04 | 81.88 | 25.52 | 76.64 | 17.58 |
| HoloGAN [33] | 34.58 | 29.81 | 44.83 | 31.25 | 52.65 | 40.70 | 48.11 | 24.17 |
| Liao et al. [24] | 17.98 | 44.29 | 39.74 | 45.70 | 33.81 | 76.42 | 31.42 | 39.09 |

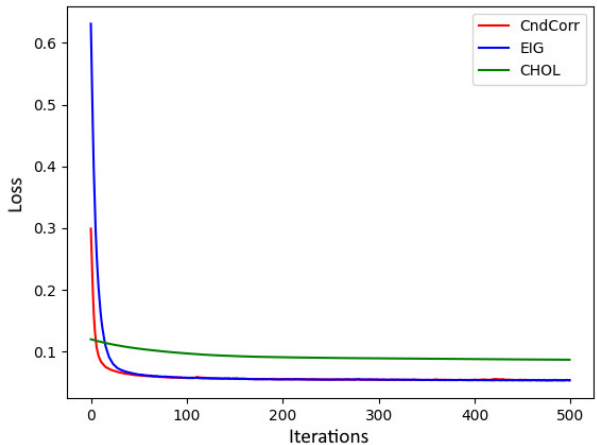

Figure 10. Average $\mathcal{L}_g$ function of iterations computed for three covariance estimation methods. While Cndcorr: conditonal correlation, and EIG: eigenvalue decomposition methods reach the same minima, CHOL: Cholesky based methods struggle due to the lack of precision propagated in the 2D projection phase. Our eigenvalue decomposition method is only slightly slower than the conditional covariance method, but it reaches the same minima and offers more intuitive control of the resulting Gaussians.

Given these correlations, our network only needs to predict six variables: $\sigma_1, \sigma_2, \sigma_3, c_{12}, c_{13}, c_{23}$. We predict all variables directly, apart from $c_{23}$, which is predicted as a combination of $c_{12}, c_{13}$:

$$c_{23} = c_{12}c_{13} + c_{23|1}\sqrt{(1-c_{12})^2(1-c_{13})^2}, \quad (9)$$

where $c_{23|1}$ is the *partial correlation* [2]. Our network predicts $c_{23|1}$ instead of $c_{23}$ and uses it to compute $c_{23}$. This approach ensures that the resulting covariance matrix is positive definite, and similarly to the eigenvalue decomposition, results on a stable training while allowing us to directly impose bounds on individual $\sigma_i$.

To evaluate these three methods, we form a test scenario where we try to optimize the Gaussian parameters for each method with the goal to fit 10 randomly selected Giraffe silhouettes; that is, we minimize the density loss $\mathcal{L}_g$.

Figure 10 shows the averaged loss values for the three methods. We see that the conditional correlation based method is the fastest to converge, followed by our eigenvalue decomposition method. Although after few iterations there is no difference between both methods. We see that the Cholesky based method fails to minimize the reconstruction objective in the same way as other methods. As such, we chose the eigenvalue based method due to its better objective minimum and its intuitive interpretation: eigenvectors represent the direction of the Gaussians while eigenvalues represent the amplitude along each direction, and this maps well to the scale and rotation of each 'part'.

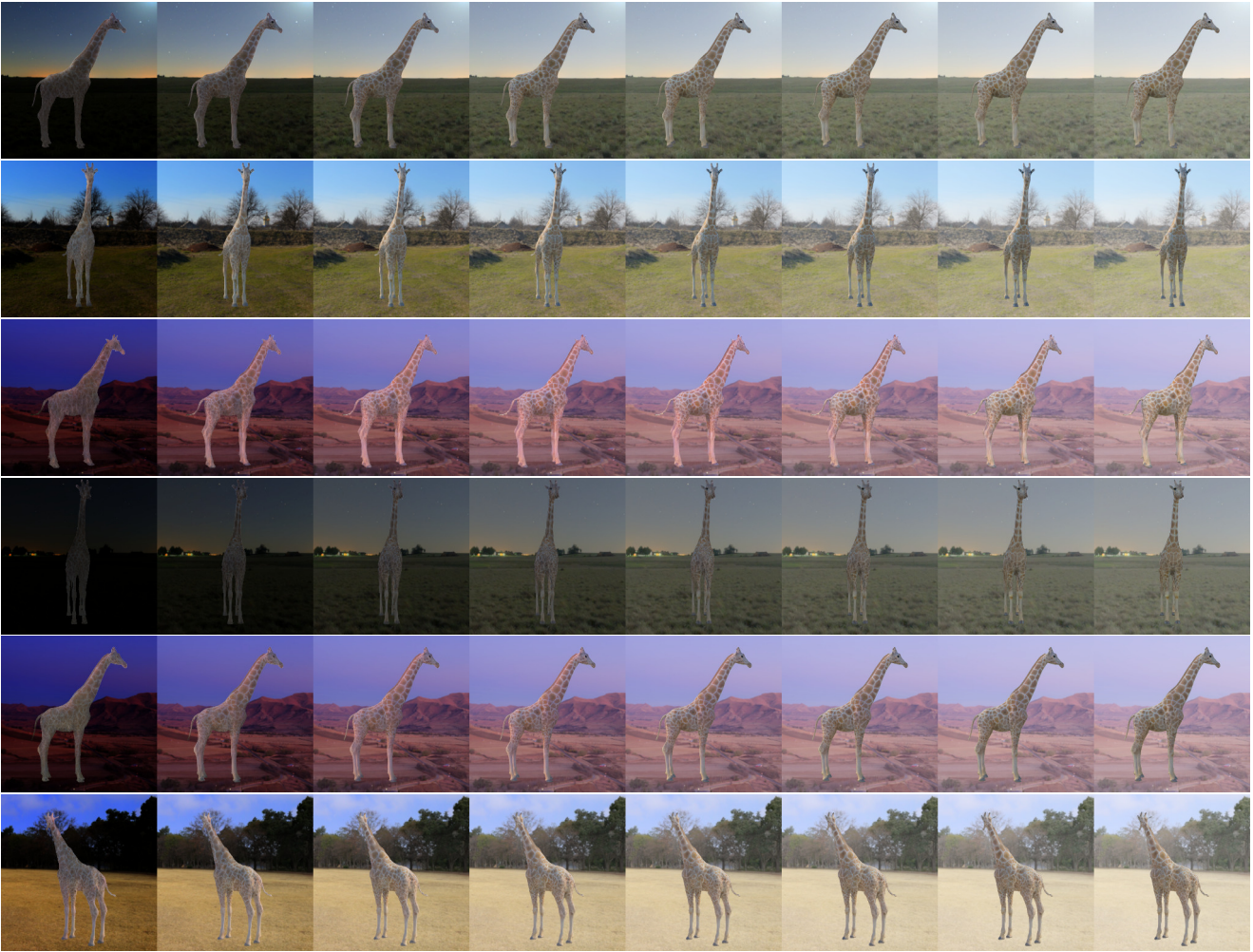

Figure 11. Lighting variation in the foreground as the background varies in intensity.

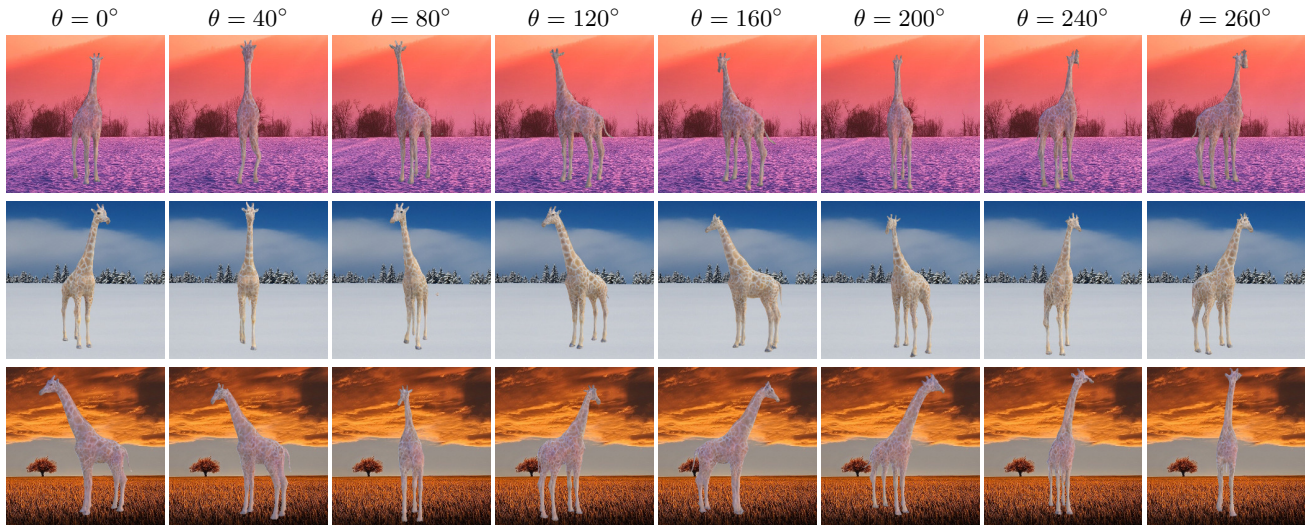

Figure 12. At test time, when giving significantly out of distribution backgrounds (e.g., fields of pink flowers under sunset, snowy landscape), our generator still produces reasonable results that match the lighting. Each row contains an example rotated from a different input instance at a different initial orientation.

## D. Additional details on mask texturing

*This section is extended from the main paper, and provides more details on the rationale for some losses.*

Given a database of RGB images $i \in \mathcal{I}$ and corresponding binary masks $m \in \mathcal{M}$, we wish to learn a generative model of texture inside the mask area conditioned on the background. For this, we compute the masked image $i_b = i \odot (1 - m)$ containing only background pixels, and the foreground image $i_f = i \odot m$, where $\odot$ is the element-wise product. We use an appearance encoder $E_i$ to extract a latent representation $z_i \in \mathbb{R}^8$ for the foreground texture: $z_i = E_i(i_f)$. The goal of $z_i$ is to supervise the texture synthesis via a separate 'latent reconstruction' loss (that we will introduce later), and to let us sample $z_i$ at test time to provide control over foreground generation.

Next, we feed the background image $i_b$ through a U-Net like architecture with residual blocks separating the encoding and decoding part. We name this network $G_i$. Through tiling, we concatenate $z_i$ layer-wise in the encoding phase of the U-Net. This helps to ensure a strong conditioning on the appearance. As an additional textural hint, we concatenate the Gaussian maps $g$ obtained from $m$ in the decoding stage of the U-net. This conditioning is also applied layer-wise in the same way $G_m$ is conditioned.

The final composited image $i'$ is thus obtained by compositing the output of $G_i$ with the original background: $i' = G_i(i_b, z_i, g) \odot m + i_b$.

**Losses.** We encourage our network to learn texture using multiple losses, with overall energy to minimize given by:

$$\mathcal{L}^t(G_i, E_i, D_{i,m}) = \beta_1 \mathcal{L}^i_{Rec} + \beta_2 \mathcal{L}^i_p + \beta_3 \mathcal{L}^i_{KL} + \beta_4 \mathcal{L}^i_{Adv} \\ + \beta_5 \mathcal{L}^i_{FM} + \beta_6 \mathcal{L}^i_{zRec}. \quad (10)$$

**Reconstruction loss.** We encourage the synthesized image $i'$ to be an identity of the input image $i$. We use the $L_1$ loss: $\mathcal{L}^i_{Rec}(i, i') = \|i - i'\|_1$'.

**Perceptual loss.** We encourage the final image to have fine-grained details by using a perceptual loss. Following Johnson *et al.* [11] we use a VGG16 network, and extract features from its second convolutional block ('conv$2_2$'), and encourage the real and generated features from $i$ and $i'$ to be similar; that is $\mathcal{L}^i_p(i, i') = \|\phi_2(i) - \phi_2(i')\|_1$, where $\phi_2$ correspond to the conv$2_2$ feature activations.

**KL loss.** To allow texture sampling of the foreground at test time, we need $z_i$ to have a constrained and known structure. Inspired from VAEs, we hence obtain $z_i$ by predicting its mean and variance vectors, then we sample it using the re-parametrization trick [17]. Using the aforementioned statistics, we enforce the latent vector to be sampled from a standard normal distribution, using the KL divergence loss.

**Latent reconstruction loss.** Even though the KL loss insures a constrained latent representation, which allows sampling from the same space encoded by $E_i$. It does not insure that the generator $G_i$ decodes different $z_i$ into diverse images. For example, $G_t$ can output the same image independently of $z_i$, and in that scenario all the losses would still be minimized. To avoid such a scenario, we add a novel encoder $E'_i$, that reconstructs the latent $z_i$ from $i'$, and enforce a reconstruction loss via: $\mathcal{L}^i_{zRec} = \|z_i - z'_i\|_1$. Note that when back-propagating gradients through that loss, we update all the parameters involved in the generation process, apart from the parameters of $E_i$. Doing so avoids the scenario where $E_i$ and $G_i$ hide the information of the latent code without producing diverse images [58].

**Adversarial loss.** We follow the same GAN loss as for the mask generation part, but the input to the new discriminator $D_{i,m}$ is now the tuple $(i, m)$, fed through concatenation. Using the tuple ensures that the generated texture is sampled from the distribution of real textures, and also that it is correlated in the same way to the mask as the real texture is. As such the adversarial loss is given by:

$$\mathcal{L}_{\text{Adv}}(G_i, D_{i,m}) = \mathbb{E}_{(i',m)}[\min(0, -D_i(i', m) - 1)] \\ + \mathbb{E}_{(i,m)}[\min(0, D_i(i, m) - 1)]. \quad (11)$$

**Feature match loss.** Separate from a perceptual loss, we also add a deep feature matching loss in a similar manner to the mask generation part. This helps improve sharpness by enforcing that real and generated images elicit similar deep feature responses:

$$\mathcal{L}^i_{\text{FM}}(D_i) = \mathbb{E}_{i,i',m}\left[\sum_{l=1}^{L}\left\|D_i^{(l)}(i', m) - \bar{D}_i^{(l)}(i, m)\right\|_2^2\right], \quad (12)$$

where $L$ is the number of feature layers within the network.

## E. Architecture details

Our network has multiple neural network based components. All the discriminators are multiscale with a depth of 3 and share the same architecture. The only variable that can change is the number of channels in the input/output layers. Please see Tables 4-9 for details on the architectures of our individual components.

**Mask generation:** Tables 4-7 show the architecture of the different components for the mask generation part.

**Texture generation:** For texture we use the same discriminator architectures used for the mask. The foreground encoder $E_i$ shares the same architecture as Table 5, but predicts mean and log-covariances that are used for sampling the texture latent using the (re)-parametrization trick, and that are also used for the KL divergence loss. Similarly, the encoder $E'_i$ that reconstructs $z_i$ from $i'$ also shares the same architecture as described in Table 5. The texture generation network is based on a U-Net like architecture (Table 9).

Table 4. Architecture for the canonical prediction network $\mathcal{G}^c$. FC refers to a fully connected layer. CONST refers to the input learnable constant. $K$ is the number of Gaussians. Blue rows correspond to the prediction of the mean vector, orange rows correspond to the prediction of the covariance matrix, and non-colored rows are shared between both.

| Layer | #neurons | Act. |
|---|---|---|
| CONST. | 256 | - |
| FC. | 256 | LReLU |
| FC. | 256 | LReLU |
| FC. | 256 | LReLU |
| FC. | 256 | LReLU |
| FC ($\boldsymbol{\mu}_k^c$). | $K*3$ | Tanh |
| FC ($\boldsymbol{v_1}$). | $K*3$ | Sigmoid |
| FC ($\boldsymbol{v_2'}$). | $K*3$ | Sigmoid |
| FC ($\boldsymbol{U}$). | $K*3$ | Sigmoid |

Table 5. Architecture for the encoder $E_m$. 'Conv.' is convolutional layer; 'Res.' is residual block; 'InstNorm' is instance normalization; 'Act.' is activation function. 'LReLU' denotes Leaky ReLU with a factor of 0.2.

| Layer | #Filters | Size | Stride | InstNorm | Act. |
|---|---|---|---|---|---|
| Conv. | 64 | $7 \times 7$ | 1 | ✓ | LReLU |
| Conv. | 64 | $3 \times 3$ | 2 | ✓ | LReLU |
| Conv. | 128 | $3 \times 3$ | 1 | ✓ | LReLU |
| Conv. | 128 | $3 \times 3$ | 2 | ✓ | LReLU |
| Conv. | 128 | $3 \times 3$ | 1 | ✓ | LReLU |
| Conv. | 128 | $3 \times 3$ | 2 | ✓ | LReLU |
| Conv. | 512 | $3 \times 3$ | 2 | ✓ | LReLU |
| MaxPool. | - | - | - | - | - |
| FC. | 8 | - | - | - | NA |

Table 6. Architecture for per-instance transforms prediction. FC refers to a fully connected layer. $K$ is the number of Gaussians. Colored rows correspond to heads for specific transforms, while uncolored rows represent the shared part of the network.

| Layer | #neurons | Act. |
|---|---|---|
| FC. | 256 | LReLU |
| FC. | 256 | LReLU |
| FC. | 256 | LReLU |
| FC. | 256 | LReLU |
| FC. | 256 | LReLU |
| FC. | 256 | LReLU |
| FC. ($\boldsymbol{t}$) | $K*3$ | Tanh |
| FC. | 256 | LReLU |
| FC. | 256 | LReLU |
| FC. ($\boldsymbol{s}$) | $K*3$ | Sigmoid |
| FC. | 256 | LReLU |
| FC. | 256 | LReLU |
| FC. ($\boldsymbol{\theta}$) | $K*3$ | Tanh |
| FC. | 256 | LReLU |
| FC. | 256 | LReLU |
| FC. ($\boldsymbol{T_0}$) | 1 | Tanh |

Table 7. Architecture for the generator $G_m$. 'T-conv.' is a transposed convolutional layer; 'InstNorm' is instance normalization; 'Act.' is activation function. 'LReLU' denotes Leaky ReLU with a factor of 0.2.

| Layer | #Filters | Size | Stride | InstNorm | Act. |
|---|---|---|---|---|---|
| T-conv. | 256 | $3 \times 3$ | 1 | ✓ | LReLU |
| T-conv. | 256 | $3 \times 3$ | 2 | ✓ | LReLU |
| T-conv. | 128 | $3 \times 3$ | 1 | ✓ | LReLU |
| T-conv. | 128 | $3 \times 3$ | 2 | ✓ | LReLU |
| T-conv. | 64 | $3 \times 3$ | 1 | ✓ | LReLU |
| T-conv. | 64 | $3 \times 3$ | 2 | ✓ | LReLU |
| Conv. | 1 | $3 \times 3$ | 1 | ✓ | Tanh |

Table 8. Architecture of the discriminators $D_m$ and $D_i$. 'LReLU' denotes Leaky ReLU with a factor of 0.2.

| Layer | #Filters | Size | Stride | InstNorm | Act. |
|---|---|---|---|---|---|
| Conv. | 64 | $4 \times 4$ | 2 | - | LReLU |
| Conv. | 128 | $4 \times 4$ | 2 | ✓ | LReLU |
| Conv. | 256 | $4 \times 4$ | 2 | ✓ | LReLU |
| Conv. | 512 | $4 \times 4$ | 1 | ✓ | LReLU |
| Conv. | 1 | $4 \times 4$ | 1 | - | Ident |

Table 9. Architecture for the texture generator $G_i$. 'Conv.' is convolutional layer; 'Res.' is residual block; 'InstNorm' is instance normalization; 'Act.' is activation function. 'LReLU' denotes Leaky ReLU with a factor of 0.2.

| Layer | #Filters | Size | Stride | InstNorm | Act. |
|---|---|---|---|---|---|
| Conv. | 64 | $7 \times 7$ | 1 | ✓ | LReLU |
| Conv. | 128 | $3 \times 3$ | 2 | ✓ | LReLU |
| Conv. | 256 | $3 \times 3$ | 2 | ✓ | LReLU |
| Conv. | 512 | $3 \times 3$ | 2 | ✓ | LReLU |
| Res. | 256 | $3 \times 3$ | 1 | ✓ | LReLU |
| Res. | 256 | $3 \times 3$ | 1 | ✓ | LReLU |
| Res. | 256 | $3 \times 3$ | 1 | ✓ | LReLU |
| Res. | 256 | $3 \times 3$ | 1 | ✓ | LReLU |
| Res. | 256 | $3 \times 3$ | 1 | ✓ | LReLU |
| Res. | 256 | $3 \times 3$ | 1 | ✓ | LReLU |
| Res. | 256 | $3 \times 3$ | 1 | ✓ | LReLU |
| Res. | 256 | $3 \times 3$ | 1 | ✓ | LReLU |
| Res. | 256 | $3 \times 3$ | 1 | ✓ | LReLU |
| Deconv. | 256 | $3 \times 3$ | 2 | ✓ | LReLU |
| Deconv. | 128 | $3 \times 3$ | 2 | ✓ | LReLU |
| Deconv. | 64 | $3 \times 3$ | 2 | ✓ | LReLU |
| Conv. | 3 | $7 \times 7$ | 1 | ✓ | LReLU |

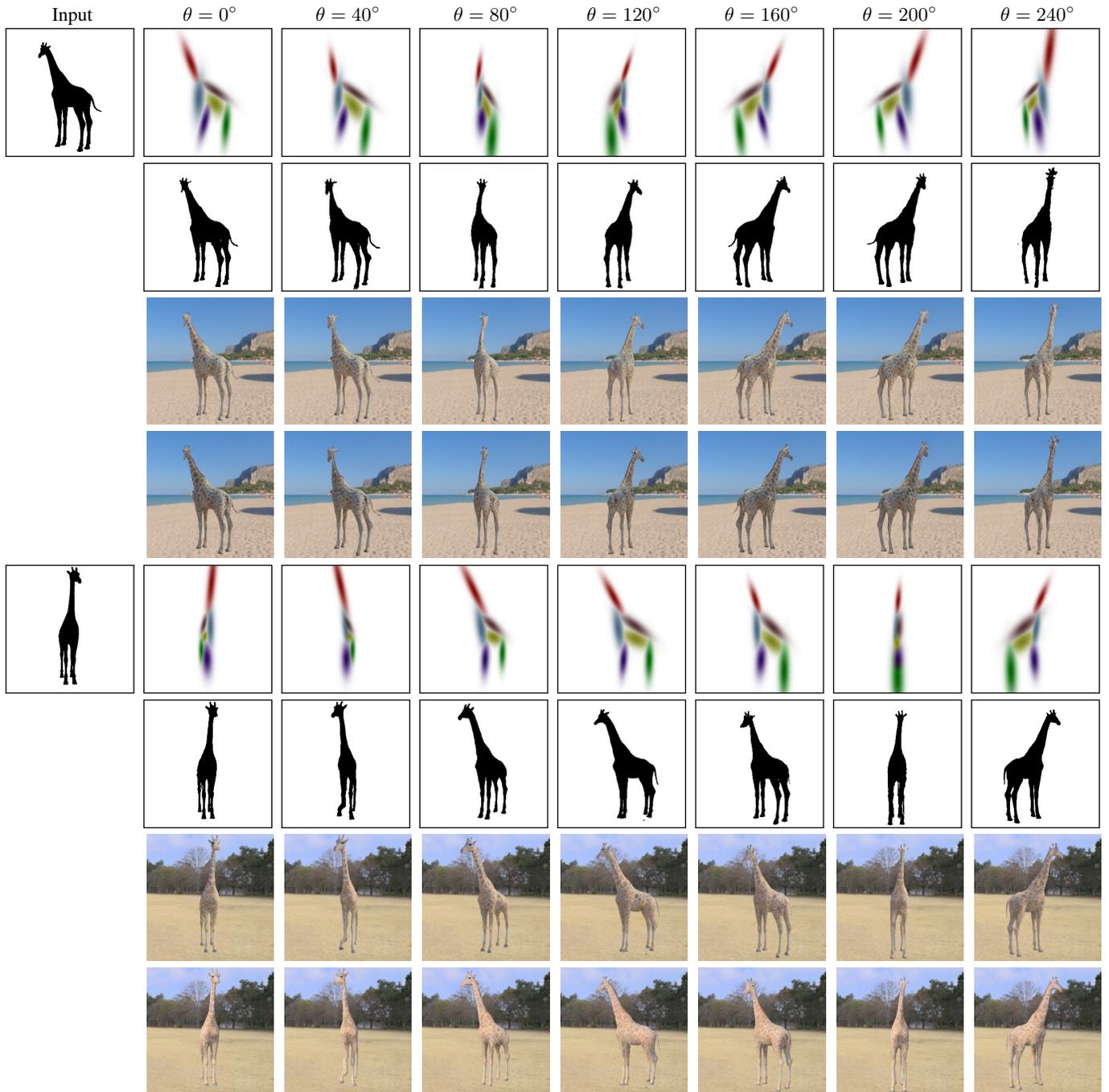

Figure 13. Additional results for Giraffe, with two randomly sampled latent vectors for texture.

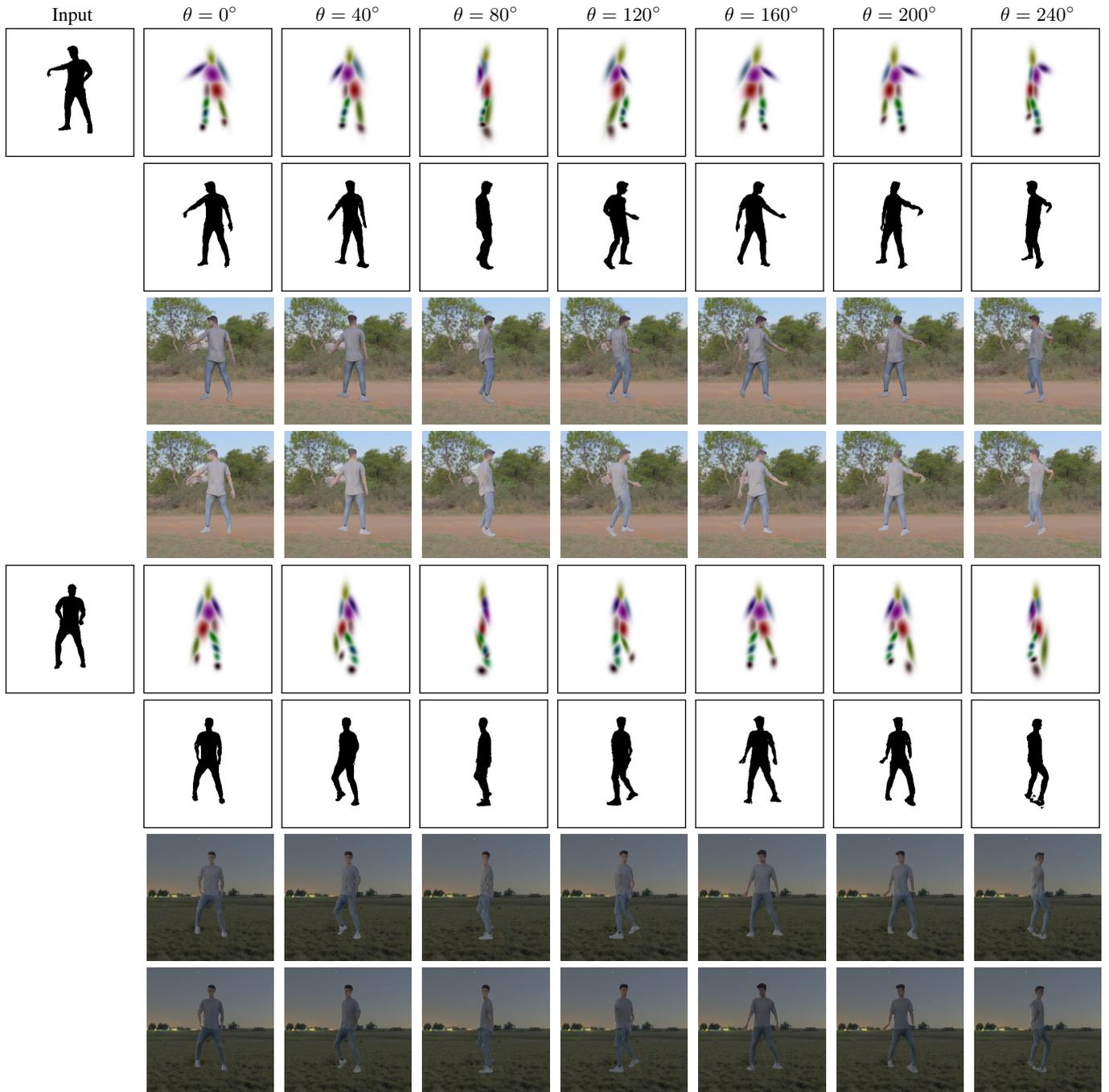

Figure 14. Additional results for Manuel, with two randomly sampled latent vectors for texture.

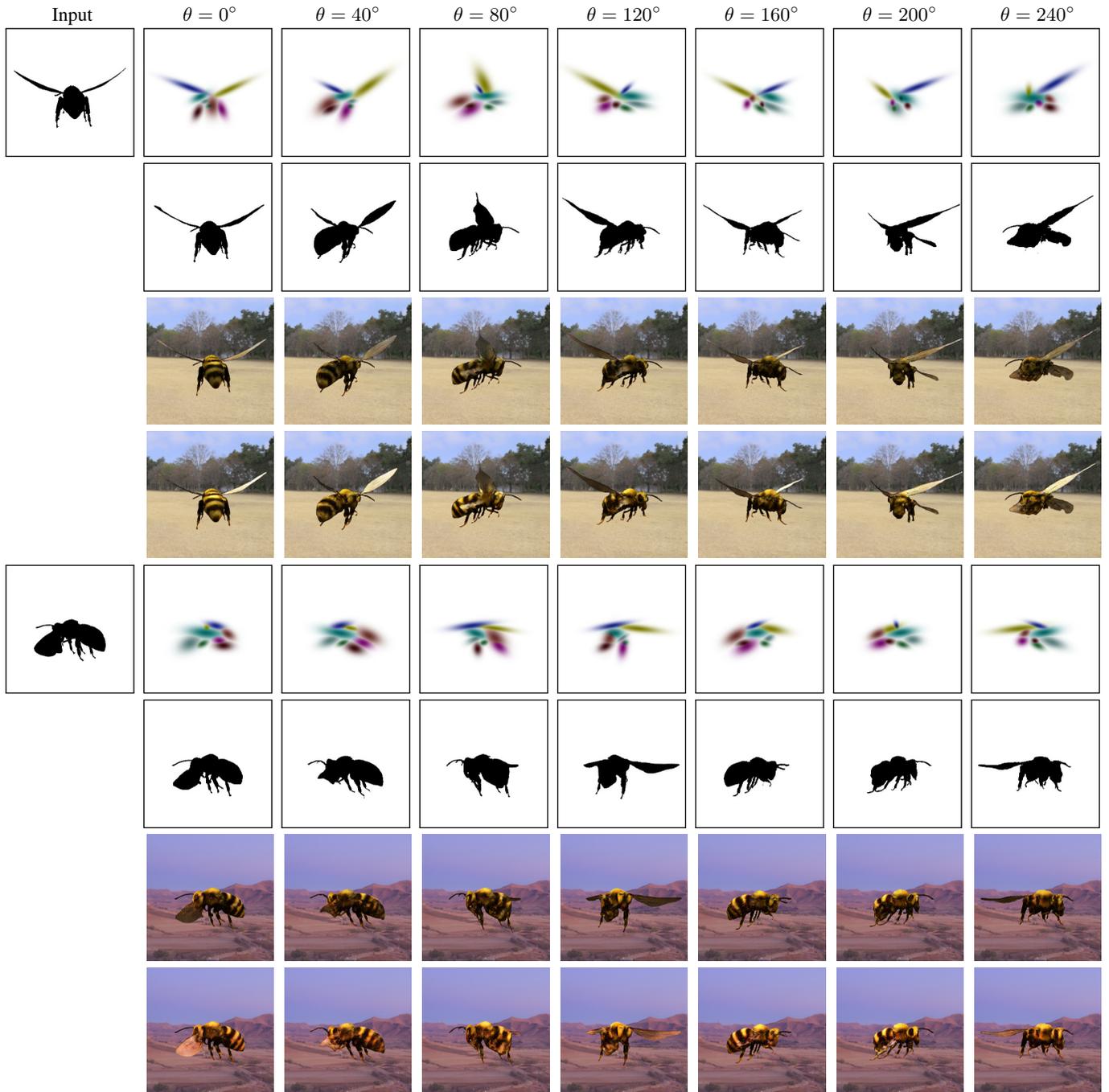

Figure 15. Additional results for beehover, with two randomly sampled latent vectors for texture.

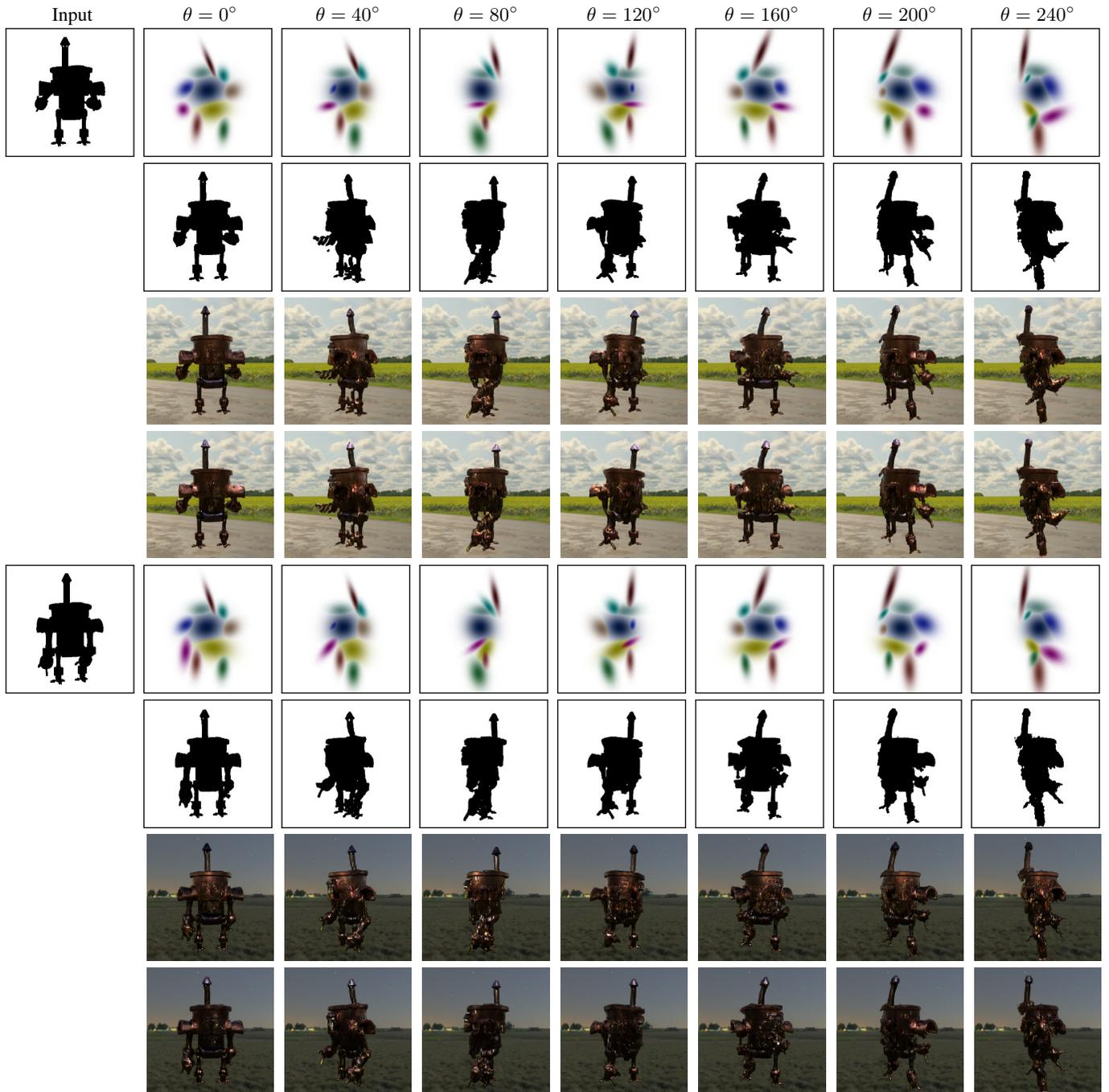

Figure 16. Additional results for oldrobot, with two randomly sampled latent vectors for texture.

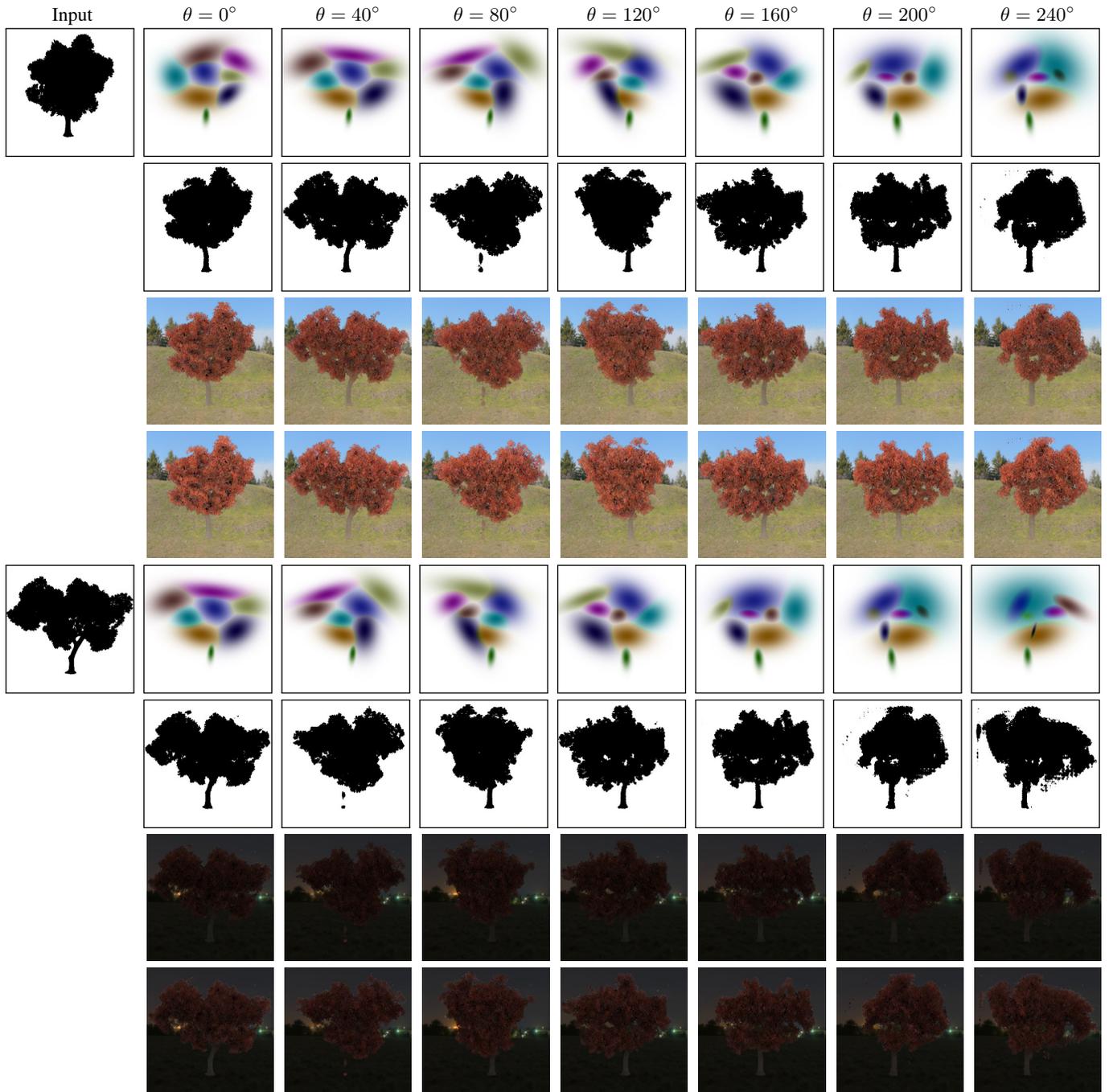

Figure 17. Additional results for Maple, with two randomly sampled latent vectors for texture.

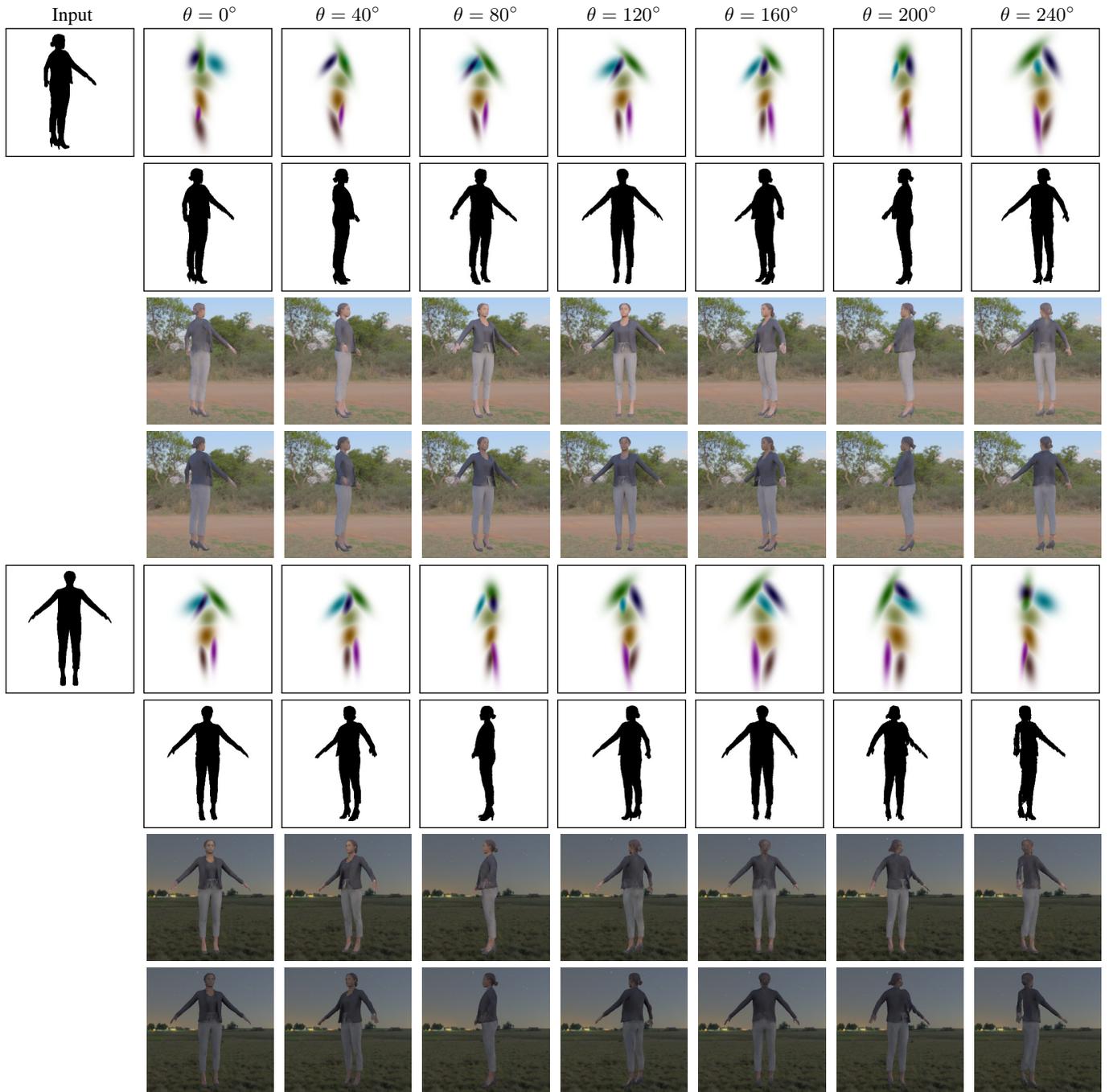

Figure 18. Additional results for Carla, with two randomly sampled latent vectors for texture.

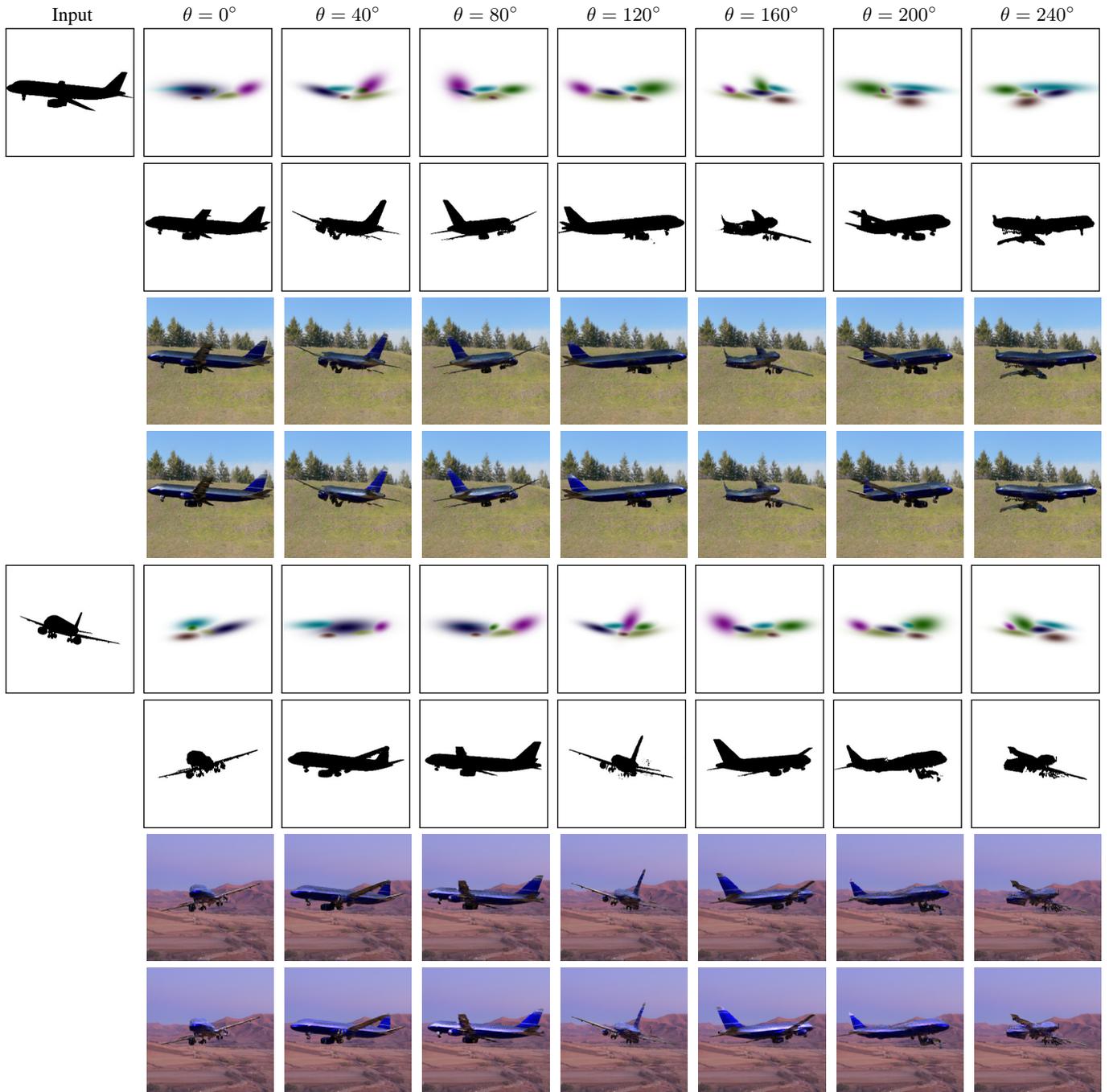

Figure 19. Additional results for airplane, with two randomly sampled latent vectors for texture.

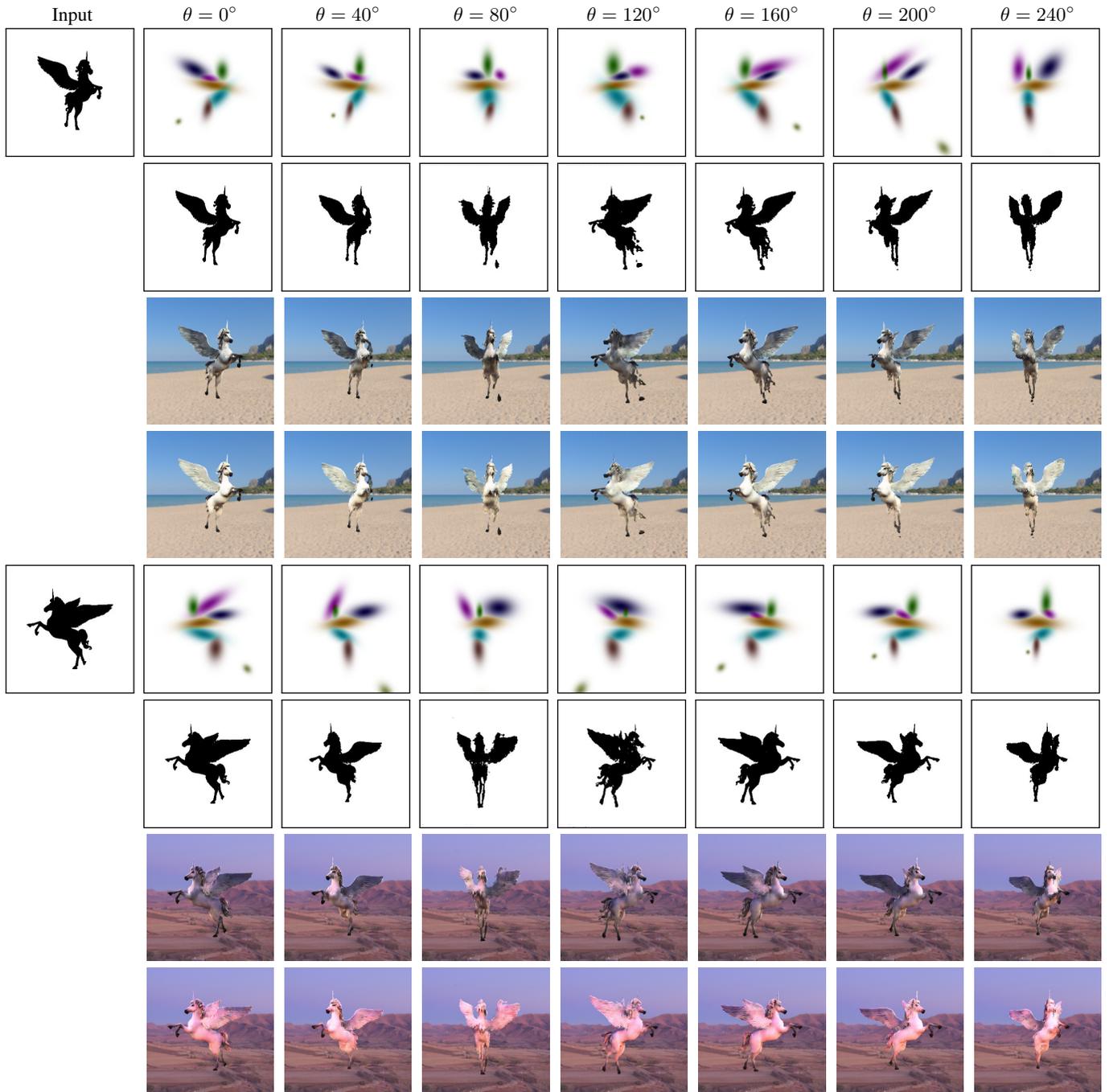

Figure 20. Additional results for pegasus, with two randomly sampled latent vectors for texture.

# F. Analytic projection of anisotropic 3D Gaussians to 2D Gaussians

Here, for reference and completeness, we reproduce the projection function $\pi$ from the supplemental material of Sridhar et al. [46] within our setting. In the main paper, we declare a general perspective pinhole camera with intrinsic matrix $\mathbf{K}$, rotation $\mathbf{R}$, and translation $\mathbf{t}$ such that camera matrix $\mathbf{P}$ is represented as $\mathbf{K}[\mathbf{R}, \mathbf{t}]$. We also declare $K$ of unnormalized anisotropic 3D Gaussians $\{\mathcal{G}_k\}_{k=1}^K$. Each Gaussian $\mathcal{G}_k$ has mean vector $\boldsymbol{\mu}_k \in \mathbb{R}^3$ and covariance matrix $\boldsymbol{\Sigma}_k \in \mathbb{R}^{3\times 3}$.

The extrinsic camera parameters are the orientation $\mathbf{R}_\phi$ of the camera at position $\boldsymbol{o}$. They transform each Gaussian $(\boldsymbol{\mu}, \boldsymbol{\Sigma})$ to the camera coordinate system by

$$\boldsymbol{\Sigma}_o = \mathbf{R}_\phi \boldsymbol{\Sigma}_k \mathbf{R}_\phi^\top, \tag{13}$$

$$\boldsymbol{\mu}_o = \mathbf{R}_\phi (\boldsymbol{\mu}_k - \boldsymbol{o}), \tag{14}$$

where the camera is looking down the positive $z$ axis. In our case, the camera's position when $\mathbf{R}_\phi = \mathbf{I}$ is at $z = 2$, with the object scaled in size to approximately fill the vertical view of the frame under $\mathbf{K}$ with angle of view equal to $90°$.

Sridhar et al. form a mathematical expression for the cone formed by rays drawn from $\boldsymbol{o}$ that are tangent to the anisotropic Gaussian. All points on this cone satisfy

$$\mathbf{x}^\top \mathbf{M} \mathbf{x} = 0, \tag{15}$$

where the cone matrix $\mathbf{M}$ is

$$\mathbf{M} = \boldsymbol{\Sigma}_o^{-1}(\boldsymbol{\mu}_o - \boldsymbol{o})\boldsymbol{\mu}_o^\top \boldsymbol{\Sigma}_o^{-1} - (\boldsymbol{\mu}_o^\top \boldsymbol{\Sigma}_o^{-1} \boldsymbol{\mu}_o - 1) \boldsymbol{\Sigma}_o^{-1}. \tag{16}$$

Points that form a projected ellipsoid on the canonical[1] image plane at $z = 1$ must also satisfy Eq. 15. Sridhar et al. derive an expression for this intersection, based on the matrix form of the second-degree polynomial representation of a conic section

$$px^2 + qxy + ry^2 + sx + ty + u = 0; \tag{17}$$

where $\mathbf{x} = [x; y; 1]^\top$. This is equivalent to Eq. 15 with $\mathbf{M}$ as

$$\mathbf{M} = \begin{bmatrix} p & q/2 & s/2 \\ q/2 & r & t/2 \\ s/2 & t/2 & u \end{bmatrix}. \tag{18}$$

Let $\mathbf{M}_{ij}$ denote the $2{\times}2$ submatrix excluding the $i$th row and $j$th column. The canonical parameters of the projected ellipse are given by

$$\widetilde{\boldsymbol{\mu}}^\pi = \frac{1}{4pr - q^2} \begin{bmatrix} qt - 2rs \\ sq - 2pt \end{bmatrix} = \frac{1}{|\mathbf{M}_{33}|} \begin{bmatrix} |\mathbf{M}_{31}| \\ -|\mathbf{M}_{23}| \end{bmatrix}, \tag{19}$$

$$\widetilde{\boldsymbol{\Sigma}}^\pi = -\frac{|\mathbf{M}|}{|\mathbf{M}_{33}|} \mathbf{M}_{33}^{-1} \tag{20}$$

For a general camera with intrinsic matrix $\mathbf{K}$, the projected ellipse $(\boldsymbol{\mu}^p, \boldsymbol{\Sigma}^p)$ from the canonical image plane is

---
[1] Unrelated to the canonical object model.

transformed to a general image plane. The transformed ellipse parameters are

$$\boldsymbol{\mu}^\pi = \mathbf{K}_{33} \widetilde{\boldsymbol{\mu}}^\pi + [k_{13}, k_{23}]^\top \tag{21}$$

$$\boldsymbol{\Sigma}^\pi = \mathbf{K}_{33} \widetilde{\boldsymbol{\Sigma}}^\pi \mathbf{K}_{33}^\top, \tag{22}$$

where $k_{ij}$ here is the entry $(i, j)$ within the $\mathbf{K}$ matrix. These equations form our 3D space and projection model.

## F.1. Camera discussion

A perspective camera was important to induce a consistent 3D space. As focal length dominates $\mathbf{K}$, and as it induces 'zooming,' one might think that perspective effects could be handled by rescaling and centering all images and masks. This may work for simpler 'sphere-like' objects or orthographic data. However, our object shape complexity, such as the long and angled neck of the giraffe, induces perspective variation under rotation, and a simpler camera model failed to learn a smooth 3D camera and object space.